\documentclass[sigconf]{acmart}
\copyrightyear{2026}
\acmYear{2026}
\setcopyright{cc}
\setcctype{by}
\acmConference[MM '26]{Proceedings of the 34th ACM International Conference on Multimedia}{November 10--14, 2026}{Rio de Janeiro, Brazil}
\acmBooktitle{Proceedings of the 34th ACM International Conference on Multimedia (MM '26), November 10--14, 2026, Rio de Janeiro, Brazil}
\acmDOI{10.1145/3767308.3835422}
\acmISBN{979-8-4007-2213-4/2026/11}
\usepackage{booktabs}
\usepackage{xcolor}
\usepackage{graphicx}
\usepackage{enumitem}
\usepackage{subcaption}
\usepackage{amsmath}
\usepackage{pifont}
\usepackage{balance}

\begin{document}

\title{VISTA: Vision-Language Integrated Semantic Triage Architecture for Online Face Clustering in Long Videos}

\author{Jiashu Liao}
\orcid{0000-0002-1058-1641}
\affiliation{%
  \institution{University of Glasgow}
  \city{Glasgow}
  \country{United Kingdom}}
\email{jiashu.liao@glasgow.ac.uk}

\author{Joanne Garde-Hansen}
\orcid{0000-0003-2462-3790}
\affiliation{%
  \institution{University of Leeds}
  \city{Leeds}
  \country{United Kingdom}}
\email{j.garde-hansen@leeds.ac.uk}

\author{Sanjay Sharma}
\orcid{0000-0003-2898-9925}
\affiliation{%
  \institution{The University of Warwick}
  \city{Coventry}
  \country{United Kingdom}}
\email{sanjay.sharma@warwick.ac.uk}

\author{Tanaya Guha}
\orcid{0000-0003-2167-4891}
\affiliation{%
  \institution{University of Glasgow}
  \city{Glasgow}
  \country{United Kingdom}}
\email{tanaya.guha@glasgow.ac.uk}

\renewcommand{\shortauthors}{Jiashu Liao, Joanne Garde-Hansen, Sanjay Sharma, and Tanaya Guha}
\begin{abstract}
Clustering faces by identity in long, continuous video streams is a challenging task that serves as a building block for human-centric video analysis. The majority of existing face clustering models are offline, i.e., they assume the entire data to be available altogether. In many practical scenarios, however, the entire data is either unavailable or poses computational challenges to be processed in its entirety. A handful of online methods that exist rely on geometric visual features and simple thresholding techniques, which fail when the faces exhibit significant appearance variations in long, unconstrained videos. To this end, we present VISTA, the first vision language framework for online face clustering in long videos. VISTA leverages semantic reasoning in a controlled, uncertainty aware fashion to maintain both high cluster purity and long range identity continuity. This makes it robust to the variations in illumination, pose, and costume changes that the faces and characters exhibit in long real world videos. Extensive experiments on the VPCD benchmark dataset (with 15 full episodes from \textit{The Big Bang Theory} [2007-2019], \textit{Buffy the Vampire Slayer} [1997-2003], and \textit{Sherlock} [2010-2017]) demonstrate that VISTA achieves a robust precision and recall balance,  consistently improving over the strongest matched online unsupervised baseline (COFC) under identical detector, embeddings, and evaluation protocol. \textbf{Code and supplementary material:} \url{https://github.com/SkyJiashu/VISTA}.
\end{abstract}



\maketitle

\section{Introduction}

\begin{figure*}[t]
\centering
\includegraphics[width=\textwidth]{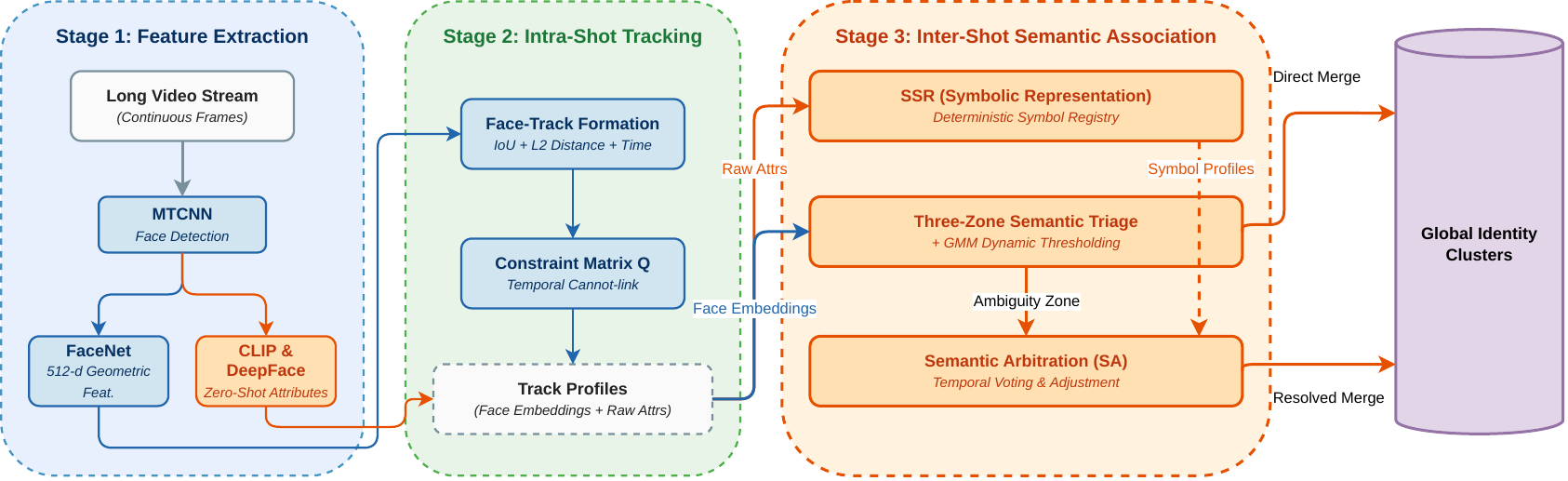} %
\caption{The VISTA pipeline. \textbf{Stage 1} extracts geometric and VLM features. \textbf{Stage 2} forms intra shot tracks. \textbf{Stage 3} performs inter shot association via SSR mapping and dynamic three zone semantic triage.}
\label{fig:pipeline}
\end{figure*}

The explosive growth of long form video content demands robust algorithms to facilitate a number of human centric video understanding tasks~\cite{wu2021towards}, such as automated character identification and narrative analysis~\cite{somandepalli2021computational, Brown2021, Haurilet2016} and personalised video browsing ~\cite{khurshid2025tvface}. At the core of these applications lies video face clustering i.e., grouping faces detected over a video stream based on identity, where each face cluster represents a unique person. The majority of existing video face clustering methods are \emph{offline} methods i.e., they process all data points simultaneously through global optimization~\cite{Brown2021,Tapaswi2019,Walawalkar2024}; however, the scale of modern video streams demands online, single pass processing algorithms that can maintain low clustering latency and remain compatible with streaming.

Strictly online face clustering algorithms are  scarce in the literature due to the constraints of single pass, low latency processing. To the best of our knowledge, Constrained Online Face Clustering (COFC)~\cite{Kulshreshtha2018} remains the state of the art for online, unsupervised face clustering. It operates by generating them incrementally using geometric embeddings (e.g., FaceNet~\cite{Schroff2015}) and spatio temporal must link/cannot link constraints. Despite the efficiency, such pure visual methods fail under extreme variations in pose and illumination, thereby often confusing identities. Several works have attempted to address this via improved face representations~\cite{Deng2019} or graph based linkage technique~\cite{Wang2019}, but the fundamental limitation is still relying solely on visual face similarity.

Recently, Vision Language Models (VLMs) like CLIP~\cite{Radford2021} have been successful in improving offline clustering~\cite{Shen2023} by extracting zero shot attributes (e.g., gender, age, clothing). However, applying VLM attributes directly as hard constraints introduces a critical \textbf{prediction instability} problem; i.e., in unconstrained videos, large variation in illumination causes VLM `hallucinations'~\cite{li2023evaluating} (e.g., a red jacket identified as black in the dark)~\cite{Shen2023}. Furthermore, as characters in long videos frequently change costumes, hard constraints based on hair or clothing will force the algorithm to reject valid faces. Because continuous temporal sequences are broken when intermediate faces are rejected, this leads to the artificial fragmentation of face trajectories. This over segmentation artificially inflates Weighted Cluster Purity (WCP) while degrading global identification accuracy~\cite{grundmann2010efficient}.

The basic idea of online face clustering is to group face tracks based on identity by processing short video segments (1 sec) as they come. If an identity has already been discovered, the new face track will be merged with that cluster, otherwise, a new cluster will be formed considering a 'new' identity. To harness VLM reasoning within this paradigm (without falling into the over segmentation trap), we propose the Vision language Integrated Semantic Triage Architecture (\textbf{VISTA}). VISTA operates as a fully unsupervised, online clustering pipeline enriched by three semantic modules. (1) First, we introduce the idea of \textbf{Symbolic Semantic Representation}: Instead of comparing raw VLM predictions, SSR maps noisy coarse age attributes (e.g., oscillating age predictions like "20-29" vs. "30-39") to a compact set of stable discrete tokens (e.g., \texttt{<age\_adult>}) via a predefined deterministic mapping table (which we term a \textit{symbol registry}), acting as a temporal denoising layer. (2) We propose a \textbf{three zone semantic triage} mechanism to improve clustering accuracy for the difficult samples. By maintaining a sliding window of the most recent 150 geometric similarity scores between track cluster pairs, we dynamically fit a Gaussian Mixture Model (GMM)~\cite{reynolds2009gaussian} to partition these track to cluster association candidates into three zones: high confidence (merge), reject, and ambiguity zone. The ambiguity zone requires VLM arbitration. (3) To resolve the ambiguity we introduce \textbf{spatio temporal decoupled semantic arbitration}. VISTA decouples attributes into \textit{persistent} (gender, age) and \textit{transient} (clothing, hair) attributes. By applying a low confidence semantic masking mechanism during poor illumination to filter out VLM hallucinations, and a temporal decay for transient attributes across long temporal gaps (bypassing costume changes), VISTA achieves more robust global identity tracking. To summarise, our main contributions are:
\begin{itemize}[leftmargin=1.5em, itemsep=0pt, parsep=0pt, topsep=2pt]
 \item We present VISTA, an online vision language framework that fuses geometric embeddings with zero shot semantic attributes. By designing the semantic integration to be  lightweight, our core clustering module operates at over 1800 FPS (clustering module only; the end to end pipeline including detection and feature extraction runs at 7--8 FPS, Section~\ref{sec:efficiency}), closing the semantic gap between low level geometric features and human interpretable identity without compromising online efficiency.
 \item We introduce the idea of a Symbolic Semantic Representation and a Three Zone Semantic Triage scheme with GMM based dynamic thresholding to adapt to the high temporal variability in facial/personal attributes in long videos. Since COFC is already online and unsupervised, the novelty of VISTA lies specifically in this selective online semantic control layer.
 \item We identify and address the VLM induced over segmentation
problem through a spatio temporal decoupled semantic arbitration
strategy. Rather than naively enforcing semantic constraints, VISTA
uses semantics selectively to reduce cross identity confusion while
maintaining a favorable trade off between cluster purity and
identity completeness across challenging long video benchmarks.
\end{itemize}
\section{Related Work}
\subsubsection*{\bf Face and person clustering in videos}
Video face clustering has traditionally been framed as an offline task. Standard approaches extract deep CNN features and perform global optimization such as graph linkage~\cite{ZhangLuo2016} or test time finetuning~\cite{Tapaswi2019}. Beyond pure face features, person level representations incorporating body and clothing cues have been explored to disambiguate visually similar faces~\cite{Sharma2019}. Sarfraz et al.\ proposed FINCH~\cite{Sarfraz2019}, a parameter free first nearest neighbour clustering method that achieves competitive results with pre trained features. Building on this, Kalogeiton and Zisserman introduced C1C~\cite{Kalogeiton2020}, which combines first nearest neighbour clustering with spatio temporal cannot link constraints derived from the video structure, improving cluster purity. For streaming scenarios, Kulshreshtha et al.\ proposed COFC~\cite{Kulshreshtha2018}, utilizing temporal overlap within shots as a strict cannot link constraint ($Q$ matrix) to guide incremental updates; this online face clustering paradigm was later extended to construct dynamic character graphs for movie analysis~\cite{Kulshreshtha2020}.  Supervised methods such as Ball Cluster Learning (BCL)~\cite{Tapaswi2019} learn to map features into pre defined metric balls; on the VPCD benchmark, BCL results are reported under the AT (unknown number of clusters) and OC (oracle number of clusters) protocols~\cite{Brown2021}, and BCL is neither unsupervised nor streaming. Multi modal approaches like MuHPC~\cite{Brown2021} hierarchically fuse face, voice, and body features to bridge otherwise unmergeable clusters, achieving strong offline results on the VPCD benchmark but remaining inherently offline and requiring multiple modalities. More recently, VideoClusterNet~\cite{Walawalkar2024} proposed self supervised refinement of face embeddings during inference, but requires multiple passes over the data. Because recent advancements have predominantly focused on offline or supervised paradigms, COFC~\cite{Kulshreshtha2018} remains the most representative baseline for fully online, unsupervised settings. We note that existing offline methods cannot be trivially adapted to online settings: MuHPC requires global hierarchical merging across the full dataset, VideoClusterNet performs multi pass self supervised refinement, and C1C relies on a complete first-nearest-neighbour graph, all fundamentally requiring full data access. VISTA builds upon this online tracking paradigm but  expands its inter shot capability by incorporating dynamic semantic triage while remaining fully unsupervised and single pass.

\begin{figure*}[t]
\centering
\includegraphics[width=\textwidth]{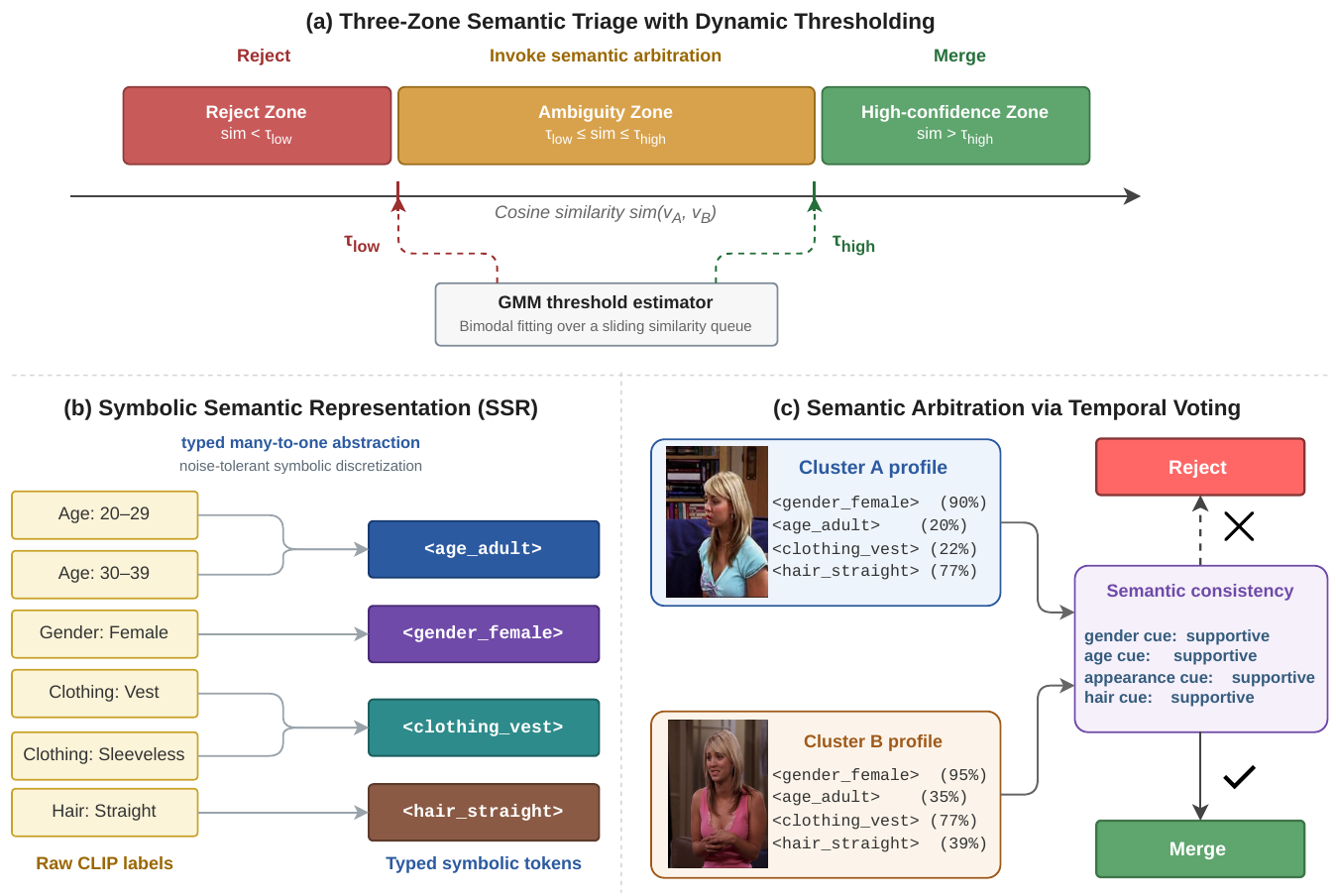}
 \caption{Detailed mechanisms of VISTA. \textbf{(a)} Three Zone Semantic Triage with GMM dynamic thresholding. \textbf{(b)} SSR maps raw predictions to stable tokens. \textbf{(c)} Semantic Arbitration via temporal voting compares symbolic profiles.}
\label{fig:mechanisms}
\end{figure*}

\subsubsection*{\bf VLMs for long video understanding}
The advent of VLMs like CLIP~\cite{Radford2021} has revolutionized zero shot video understanding. For face analysis, Shen et al.\ introduced CLIP-Cluster~\cite{Shen2023}, showing that VLMs can extract semantic attributes to assist in face grouping. Subsequent large scale vision language models such as BLIP2~\cite{Li2023} and LLaVA~\cite{Liu2024} further improved zero shot recognition capabilities, enabling richer attribute extraction from video frames. However, VLM outputs are  susceptible to motion blur and illumination drops in the wild. DeepFace~\cite{Serengil2024} offers a complementary source of demographic predictions but similarly degrades under low image quality. Earlier attempts at naming TV characters also leveraged multimodal cues from subtitles and dialog~\cite{Haurilet2016}, yet these require structured metadata not always available in general video streams. VISTA repurposes VLM predictions as a source of symbolic supervision, employing a spatio temporal decoupling mechanism to dynamically penalize or ignore semantic contradictions based on environmental context, thereby bridging the gap between powerful zero shot capabilities and the practical instability of real world deployment.

\section{Methodology}
The VISTA framework processes streaming video through three cascaded stages: (1) Feature Extraction, (2) Intra Shot Tracking, and (3) Inter Shot Semantic Association. Figure~\ref{fig:pipeline} illustrates the overall pipeline. Pseudocode for the complete online clustering procedure is provided in the Supplementary Material. Geometric embeddings and CLIP/DeepFace attributes are extracted up front for every face crop in Stage 1; triage controls only how often semantic \emph{arbitration} fires (Section~3.3).

\subsection{Problem Formulation}
Given a continuous video stream $\mathcal{V} = \{f_1, f_2, \ldots\}$, we detect face instances and seek to assign each detection to an identity cluster in an online fashion, without revisiting past frames. Let $\mathcal{D}_t = \{d_1^t, d_2^t, \ldots\}$ denote detected faces at frame $t$, each associated with a geometric embedding $v \in \mathbb{R}^{512}$ and a set of raw attribute predictions $\mathcal{A} = \{a_{\text{gender}}, a_{\text{age}}, a_{\text{clothing}}, a_{\text{hair}}\}$. The goal is to maintain a set of global identity clusters $\mathcal{C} = \{C_1, C_2, \ldots, C_K\}$ where $K$ is unknown and grows dynamically. Each incoming face track must be either merged into an existing cluster or instantiated as a new identity, using only information available up to the current time step.
\subsection{Feature Extraction and Intra Shot Tracking}
Faces are detected using RetinaFace~\cite{Deng2020RetinaFace} and represented via FaceNet \cite{Schroff2015} embeddings. While our architecture is modular and supports alternative backends (e.g., InsightFace~\cite{Guo2021InsightFace}), we employ this specific setup by default to maintain a strict, fair comparison with classical online baselines like COFC. Within a camera shot, we construct short term face tracks $T$ based on bounding box IoU and $L_2$ feature distance. Specifically, a detection $d_i^t$ is appended to an existing track $T_j$ if $\text{IoU}(d_i^t, d_j^{t-1}) > \theta_{\text{iou}}$ and $\|v_i^t - \bar{v}_j\|_2 < \theta_{\text{feat}}$, where $\bar{v}_j$ is the running average embedding of $T_j$. Each track accumulates up to $L_{\text{max}}$ frames (default 24) to form a stable representation. To enforce physical realism, we construct a temporal constraint matrix $Q$. If two tracks co-occur in the same frame, $Q_{A,B}=0$ (cannot link constraint); otherwise $Q_{A,B}=1$.

\begin{table*}[t]
\centering
\caption{ Face clustering performance. \textbf{Top:} controlled comparison, VISTA vs.\ COFC under identical detector (RetinaFace), embeddings (FaceNet), and protocol (WCP / NMI / ID\_Acc \%). \textbf{Bottom:} literature reference numbers with SENet-50 features$^\dagger$ (WCP / NMI only); feature/protocol differences make these reference points, not controlled comparisons. BCL~(AT)/(OC) denote the VPCD protocols with unknown/oracle cluster counts as reported by Brown et al.~\cite{Brown2021}; BCL is supervised and not streaming. Bold: best in controlled comparison.}
\label{tab:main_results}
\resizebox{\textwidth}{!}{
\begin{tabular}{lcccccc}
\toprule
Method & Online? & Unsup.? & Core Mechanism & \textit{TBBT} & \textit{Buffy} & \textit{Sherlock} \\
\midrule
\multicolumn{7}{l}{\textbf{~Controlled comparison: online, unsupervised, identical features and protocol (WCP / NMI / ID\_Acc)}} \\
COFC~\cite{Kulshreshtha2018} & \ding{51} & \ding{51} & Constrained Incremental & 92.13 / 86.97 / 89.74 & 64.60 / 66.28 / 64.31 & 76.22 / 50.56 / \textbf{71.27} \\
\textbf{VISTA (Full Framework)} & \ding{51} & \ding{51} & Spatio Temporal Semantic Triage & \textbf{96.97} / \textbf{91.25} / \textbf{93.94} & \textbf{87.39} / \textbf{79.30} / \textbf{82.82} & \textbf{80.45} / \textbf{51.19} / 68.10 \\
\midrule
\multicolumn{7}{l}{\textbf{~Literature reference: SENet-50 features$^\dagger$, reported numbers (WCP / NMI)}} \\
C1C$^\dagger$~\cite{Kalogeiton2020} & \ding{55} & \ding{51} & Constrained 1NN & 89.20 / 87.40 & 66.30 / 68.80 & 76.30 / 50.30 \\
\multicolumn{7}{l}{\textbf{}} \\
~BCL$^\dagger$~(AT)~\cite{Tapaswi2019,Brown2021} & \ding{55} & \ding{55} & Ball Cluster Learning & 90.80 / 85.70 & 85.00 / 78.80 & - / - \\
\multicolumn{7}{l}{\textbf{}} \\
~BCL$^\dagger$~(OC)~\cite{Tapaswi2019,Brown2021} & \ding{55} & \ding{55} & ~Ball Cluster Learning & 94.00 / 85.00 & 86.50 / 77.60 & - / - \\
MuHPC$^\dagger$ (Face)~\cite{Brown2021} & \ding{55} & \ding{55} & Global Hierarchical 1NN & 99.40 / 97.80 & 96.10 / 92.80 & 86.70 / 60.30 \\
\bottomrule
\end{tabular}}
\end{table*}
\subsection{Inter Shot Clustering: Semantic Association}
\subsubsection*{\bf Symbolic Semantic Representation (SSR)}
\label{sec:ssr}
Applying raw VLM textual predictions (e.g., `Age: 20-29') directly is fragile due to frame level noise. VISTA defines a Symbol Registry, consisting of deterministic mapping functions $\phi_d : L_d \rightarrow S_d$ that convert coarse age grouping labels into discrete symbols. For example, both `Age: 20-29' and `Age: 30-39' map to the symbol \texttt{<age\_adult>}, absorbing boundary oscillations. We define four attribute dimensions $d \in \{\text{gender}, \text{age}, \text{clothing}, \text{hair}\}$, each with a compact symbol vocabulary $|S_d| \in \{2, 5, 7, 5\}$ respectively. The exact mapping schemes and an empirical comparison between different age symbol configurations are provided in the Supplementary Material. Overall, the coarse grouping used in our unified setting provides a robust compromise across datasets by reducing boundary oscillation while retaining more discriminative power than broader adult collapsing.

Over the duration of a track, VISTA accumulates these symbols. The track's final SSR profile is generated via a temporal majority vote, acting as a robust denoising mechanism. The confidence $p(s)$ of a symbol $s$ in a track is defined as its frequency of occurrence across all valid frames in the track:
\begin{equation}
 p_T(s_d) = \frac{|\{f \in T : \phi_d(a_d^f) = s_d\}|}{|\{f \in T : a_d^f \neq \texttt{null}\}|}.
\end{equation}
Only frames where the attribute prediction passes a minimum detection quality threshold contribute to the vote, filtering out  corrupted observations.

\subsubsection*{\bf Three zone triage with GMM dynamic thresholding}
To evaluate a merge between a query track $T$ and a global cluster $C$, we compute the cosine similarity $\mathrm{sim}(v_T,v_C)$. Using thresholds $\tau_{\text{high}}$ and $\tau_{\text{low}}$, VISTA triages pairs into three zones, each activating a different \emph{mode} of semantic arbitration: (1) \textit{High confidence zone} ($\mathrm{sim} > \tau_{\text{high}}$): merge is accepted; semantic cues contribute \emph{bonus only} adjustments (no penalties) to reinforce the decision; (2) \textit{Ambiguity zone} ($\tau_{\text{low}} \le \mathrm{sim} \le \tau_{\text{high}}$): full semantic arbitration with both rewards and penalties; (3) \textit{Reject zone} ($\mathrm{sim} < \tau_{\text{low}}$): merge is rejected by default, but a \emph{rescue} mechanism can override the rejection when persistent semantic cues (gender + age) strongly agree.

Rather than relying exclusively on static boundaries, VISTA maintains a sliding window queue $\mathcal{Q}$ of recent similarity scores and fits a bimodal Gaussian Mixture Model (GMM) online to capture the distribution of same identity and different identity pairs. During the initial warm up phase (i.e., before $\mathcal{Q}$ is fully populated), the system relies on static initial thresholds $\tau_{\text{high}}^{(0)}$ and $\tau_{\text{low}}^{(0)}$. The sensitivity of these initial parameters is extensively evaluated in the Supplementary Material. The GMM is formalized as:
\begin{equation}
 p(x) = \pi_1 \mathcal{N}(x \mid \mu_1, \sigma_1^2) + \pi_2 \mathcal{N}(x \mid \mu_2, \sigma_2^2),
\end{equation}
where the lower component ($\mu_1 < \mu_2$) models different identity similarities and the upper component models same identity similarities.  We dynamically estimate the threshold proposals as
{
\begin{equation}
 \tau_{\text{low}}^\star=\mathrm{clip}(\mu_1+\sigma_1,\,0.15,\,0.50), \qquad \tau_{\text{high}}^\star=\mu_2-0.5\sigma_2.
\end{equation}}
The $0.50$ cap on $\tau_{\text{low}}^\star$ is a deliberate design choice: it preserves a sufficiently wide ambiguity zone for semantic arbitration in visually compressed windows. A cap sweep over all 15 episodes (Supplementary Material) shows that relaxing it to $0.70$ or $1.00$ leaves all metrics unchanged after rounding, i.e., it is a stability safeguard rather than a tuned knob. To ensure temporal stability, we apply an exponential moving average update~\cite{tarvainen2017mean}:
\begin{equation}
 \tau_{\text{new}} = \eta_{\text{gmm}} \tau^\star + (1-\eta_{\text{gmm}})\tau_{\text{old}},
\end{equation}
where $\eta_{\text{gmm}}=0.3$ is the GMM threshold smoothing coefficient, distinct from the cluster representation momentum $\texttt{ema\_rep}$ (Section~\ref{sec:impl}). This mechanism allows VISTA to adapt to evolving visual distributions across different scenes while preventing abrupt threshold oscillations. An empirical visualization confirming the bimodal structure of the similarity distributions across all 15 episodes is provided in the Supplementary Material.

\subsubsection*{\bf Spatio temporal decoupled arbitration}
For pairs in the ambiguity zone ($\tau_{\text{low}} \le \mathrm{sim} \le \tau_{\text{high}}$), VISTA compares their SSR profiles to compute an adjustment score $\mathrm{adj}$. We categorize attributes into persistent traits $\mathcal{P} = \{\text{gender}, \text{age}\}$ and transient traits $\mathcal{T} = \{\text{clothing}, \text{hair}\}$. We define a joint confidence weight $w=\min(p_T(s), p_C(s))$. The adjustment combines reward and penalty terms:
\begin{equation}
 \mathrm{adj} = \underbrace{\sum_{r} \lambda_r \cdot w_r}_{\text{rewards}} - \underbrace{\gamma \cdot \sum_{p} \lambda_p \cdot w_p}_{\text{penalties}},
\end{equation}
where $\gamma$ is a global penalty scaling factor. Reward terms include gender+age agreement ($\lambda_{\text{ga}}$) and clothing consistency ($\lambda_{\text{cloth}}$), while penalties include gender mismatch ($\lambda_{\text{gm}}$), age mismatch ($\lambda_{\text{am}}$), and hair conflict ($\lambda_{\text{hc}}$). The final updated similarity is $\mathrm{sim}'=\mathrm{sim}+\mathrm{adj}$. A merge is accepted if $\mathrm{sim}' \ge \tau_{\text{merge}}$. We set $\tau_{\text{merge}}=0.325$: it is the final acceptance gate for ambiguity zone pairs \emph{after} arbitration, not a standalone visual threshold; a pair must be the best current assignment and its post arbitration score, after possible penalties, must pass the gate, whereas high confidence pairs are routed directly by $\tau_{\text{high}}$. A sweep over $\{0.280$--$0.475\}$ (Supplementary Material) confirms the expected purity--continuity trade off. To prevent over segmentation, VISTA decouples the spatial and temporal dimensions as follows:\\

\noindent \emph{Low Confidence Semantic Masking.} When the mean face detection confidence $\bar{c}_T$ of a track drops below a quality threshold $\theta_{\text{cliff}}$ (indicating blur or darkness), VISTA continuously attenuates all semantic penalties via a linear scaling factor:
\begin{equation}
\gamma_{\text{eff}} = \min\!\Bigl(1.0,\; \max\!\bigl(0.1,\; \tfrac{\bar{c}_T - \theta_{\text{floor}}}{\theta_{\text{cliff}} - \theta_{\text{floor}}}\bigr)\Bigr),
\end{equation}
where $\theta_{\text{floor}}$ is a lower bound below which attenuation saturates. This soft masking ensures that as visual quality degrades, the influence of potentially hallucinated VLM predictions is progressively suppressed rather than abruptly removed, avoiding discontinuous clustering behaviour at the quality boundary.\\

\begin{table*}[t]
\centering
\caption{Ablation study of semantic components (WCP / NMI / ID\_Acc \%).}
\label{tab:ablation}
{
\begin{tabular}{lccc}
\toprule
Configuration & \textit{TBBT} & \textit{Buffy} & \textit{Sherlock} \\
\midrule
VISTA (Full Framework) & \textbf{96.97} / \textbf{91.25} / \textbf{93.94} & \textbf{87.39} / \textbf{79.30} / \textbf{82.82} & 80.45 / 51.19 / 68.10 \\
\midrule
w/o Semantic Arbitration  & 96.71 / 89.97 / 91.27 & 80.54 / 77.62 / 78.35 & 79.06 / 52.55 / 69.56 \\
w/o SSR  & 94.83 / 89.29 / 91.60 & 74.75 / 74.47 / 73.31 & 78.36 / \textbf{52.62} / \textbf{71.43} \\
w/o Three Zone Gating  & 96.78 / 90.80 / 92.32 & 80.17 / 77.03 / 76.58 & 80.27 / 51.49 / 64.63 \\
w/o Temporal Smoothing & 96.28 / 90.09 / 91.95 & 82.07 / 77.44 / 77.05 & \textbf{84.11} / 52.54 / 62.23 \\
\bottomrule
\end{tabular}}
\end{table*}

\noindent \textit{Temporal Decay for Transient Attributes.} If the frame gap $\Delta f$ between a track and a cluster exceeds a long range threshold $\Delta_{\text{max}}$, VISTA masks transient traits (e.g., clothing, hair) as \texttt{unknown}, bypassing costume driven conflicts:
\begin{equation}
 w_d^{\text{eff}} = \begin{cases} w_d & \text{if } d \in \mathcal{P} \text{ or } \Delta f \le \Delta_{\text{max}}, \\ 0 & \text{if } d \in \mathcal{T} \text{ and } \Delta f > \Delta_{\text{max}}. \end{cases}
\end{equation}
This mechanism ensures that VISTA relies more on persistent traits for long range temporal association, while still leveraging transient cues for temporally adjacent merges where costume stability can be assumed.

\subsubsection*{\bf Perplexity based confidence weighting}
To further modulate the influence of semantic arbitration, VISTA maintains a short history window of $H$ recent per cluster entropy observations. When the perplexity of a cluster's embedding distribution exceeds a threshold, indicating high intra cluster visual diversity, the semantic adjustment magnitude is down weighted by a factor
\begin{equation}
\omega = \max\bigl(\omega_{\min},\; e^{-\beta \cdot \mathrm{perp}}\bigr),
\end{equation}
where $\beta$ controls the decay rate. This prevents aggressive semantic penalties for clusters that already exhibit high visual variance.

\section{Experiments}
\subsection{Datasets and Evaluation Metrics}
We evaluate on the VPCD benchmarks: \textbf{\textit{TBBT}} (6 episodes, well lit sitcom), \textbf{\textit{Sherlock}} (3 episodes, thriller with costume changes), and \textbf{\textit{Buffy}} (6 episodes, dark action, which feature rich narrative and character arcs~\cite{kaveney2004reading}). We report \textbf{Weighted Cluster Purity (WCP)}, \textbf{Normalized Mutual Information (NMI)}, and \textbf{Identification Accuracy (ID\_Acc)}. WCP focuses on cluster homogeneity, while ID\_Acc penalizes trajectory fragmentation. VISTA seeks a favorable balance across these metrics.

\subsection{Implementation Details}
\label{sec:impl}
VISTA is implemented in Python using PyTorch. As described in Section 3.2, we utilize RetinaFace~\cite{Deng2020RetinaFace} for face detection and FaceNet~\cite{Schroff2015} for identity embedding. CLIP ViT-B/32~\cite{Radford2021} performs zero shot attribute classification with text prompts for clothing and hair, and DeepFace~\cite{Serengil2024} is used for age analysis. The complete CLIP text prompt templates, label vocabularies, and the full Symbol Registry mappings are provided in the Supplementary Material.

For intra shot tracking, we use $\theta_{\text{iou}} = 0.3$ and $\theta_{\text{feat}} = 1.0$. In the final full framework evaluation, we use a single unified configuration across all three datasets to avoid dataset specific tuning and demonstrate robust generalization: RetinaFace for detection, FaceNet for embedding, \texttt{coarse age grouping} age symbols, $\tau_{\text{high}}^{(0)}=0.85$, $\tau_{\text{low}}^{(0)}=0.25$,  cluster representation smoothing momentum $\texttt{ema\_rep}=0.95$, GMM threshold smoothing $\eta_{\text{gmm}}=0.3$, arbitration gate $\tau_{\text{merge}}=0.325$, GMM queue size $|\mathcal{Q}|=150$, perplexity weighting enabled, and fixed length scene segmentation with length 24. Dynamic thresholding and semantic arbitration are enabled. Perplexity weighting uses a history window of $H=20$ observations with $\omega_{\min}=0.10$ and $\beta=1.2$. For semantic arbitration (Eq.~5), the adjustment weights are empirically set as $\lambda_{ga}=0.08$, $\lambda_{cloth}=0.04$, $\lambda_{gm}=0.12$, $\lambda_{am}=0.02$, and $\lambda_{hc}=0.04$. A grid search over 243 $\lambda$ configurations (detailed in the Supplementary Material) shows that VISTA's performance is stable across a wide range of settings: WCP standard deviation is only $0.31\%$ on \textit{TBBT}, $1.09\%$ on \textit{Buffy}, and $0.78\%$ on \textit{Sherlock}, confirming that the method is not driven by fragile hyperparameter tuning.

To study robustness, we additionally perform a parameter sweep over $\tau_{\text{high}}^{(0)}$, $\tau_{\text{low}}^{(0)}$, age symbol configuration, $\texttt{ema\_rep}$ momentum, and GMM queue size. The detailed dataset specific parameter effect tables are reported in the Supplementary Material and should be interpreted as marginal robustness studies rather than the source of the final unified full framework setting; key trends there: coarse age grouping and strong $\texttt{ema\_rep}$ smoothing yield consistently best results, and darker scenes (\textit{Buffy}) favor larger GMM queues for stable threshold estimation. In addition, a $3\times3$ grid bracketing the warm up initialisation over $\tau_{\text{high}}^{(0)}\in\{0.75, 0.825, 0.90\}$ and $\tau_{\text{low}}^{(0)}\in\{0.20, 0.275, 0.35\}$ (135 episode level evaluations, all else fixed) yields max$-$min variation of at most $0.69$\,pp across all metrics and datasets (Supplementary Material); $\tau_{\text{high}}$ sensitivity is limited precisely because GMM dynamic thresholding overrides the static initialisation after warm up.

All experiments are conducted on a single NVIDIA RTX A6000 GPU. A detailed breakdown of the computational efficiency and throughput of our pipeline is discussed in Section \ref{sec:efficiency}.
\subsection{Quantitative Results}
 Table~\ref{tab:main_results} reports a controlled comparison between VISTA and COFC under identical detector, embeddings, and protocol, together with literature reported results as reference points. We note that strictly online unsupervised face clustering methods are rare; COFC~\cite{Kulshreshtha2018} remains the only published method in this category with results on VPCD. We additionally include offline unsupervised (C1C~\cite{Kalogeiton2020}), supervised (BCL~\cite{Tapaswi2019}), and multi-modal (MuHPC~\cite{Brown2021}) methods as reference points. Methods marked with $\dagger$ use SENet-50 features, which are generally more discriminative than the FaceNet embeddings used by VISTA and COFC~\cite{Deng2019}.

VISTA outperforms COFC across all three datasets under identical features and evaluation. On \textit{TBBT}, WCP and NMI improve from $92.13\%$/$86.97\%$ to $96.97\%$/$91.25\%$, and ID\_Acc from $89.74\%$ to $93.94\%$. On \textit{Buffy}, VISTA raises WCP and NMI from $64.60\%$/$66.28\%$ to $87.39\%$/$79.30\%$, with ID\_Acc improving from $64.31\%$ to $82.82\%$ ($+18.51$\,pp), also far surpassing the offline C1C ($66.30\%$/$68.80\%$) that uses stronger features.  As literature reference points, VISTA's \textit{TBBT} results ($96.97\%$/$91.25\%$) are also higher than the reported supervised BCL~(AT) numbers ($90.80\%$/$85.70\%$), and its \textit{Buffy} NMI is comparable ($79.30\%$ vs.\ $78.80\%$), although feature and protocol differences preclude a strict comparison.

On \textit{Sherlock}, VISTA improves over both COFC ($76.22\%$/$50.56\%$) and C1C ($76.30\%$/$50.30\%$), reaching $80.45\%$/$51.19\%$. However, ID\_Acc decreases from $71.27\%$ (COFC) to $68.10\%$ ($-3.17$\,pp): under \textit{Sherlock}'s heavy disguises and long gap appearance variation, VISTA prioritises cluster purity over identity continuity, yielding cleaner clusters at the cost of additional fragmentation. We regard this as a limitation rather than a universal all metric gain; Sections~\ref{sec:ablation} and~\ref{sec:qualitative} analyse this purity--continuity trade off in detail. Although the gains are more modest than on \textit{TBBT} and \textit{Buffy}, this confirms that selective semantic integration provides consistent WCP/NMI improvements even on the most volatile dataset, with room for future refinement.

\paragraph{Additional online baselines.} To isolate the source of the gains, we evaluate two further online baselines under the same features and protocol (full tables in the Supplementary Material). (i) \emph{Naive VLM hard constraints}: raw attributes applied as must link/cannot link rules (no SSR, no triage) over fragment severely, producing $4.0$--$11.3\times$ more clusters than ground truth; the inflated cluster count artificially raises WCP while ID\_Acc collapses to $63.95$/$60.64$/$40.65$ on \textit{TBBT}/\textit{Buffy}/\textit{Sherlock}, far below VISTA Full and below our weakest geometry driven ablation (Table~\ref{tab:ablation}). This confirms that VISTA's improvements stem from \emph{selective semantic triage}, not from VLM attributes alone. (ii) \emph{Sequential online HAC} attains very high WCP ($99.73$/\allowbreak$97.62$/\allowbreak$91.19$) under severe fragmentation (\#Pred/\#GT $=4.5$/\allowbreak$9.5$/\allowbreak$39.8\times$) but substantially lower NMI/ID\_Acc ($84.93$/\allowbreak$81.29$, $76.78$/\allowbreak$63.49$, $49.28$/\allowbreak$40.56$), illustrating the over segmentation trap that three zone triage is designed to avoid.

\begin{figure*}[!t]
    \centering
    \includegraphics[width=\textwidth]{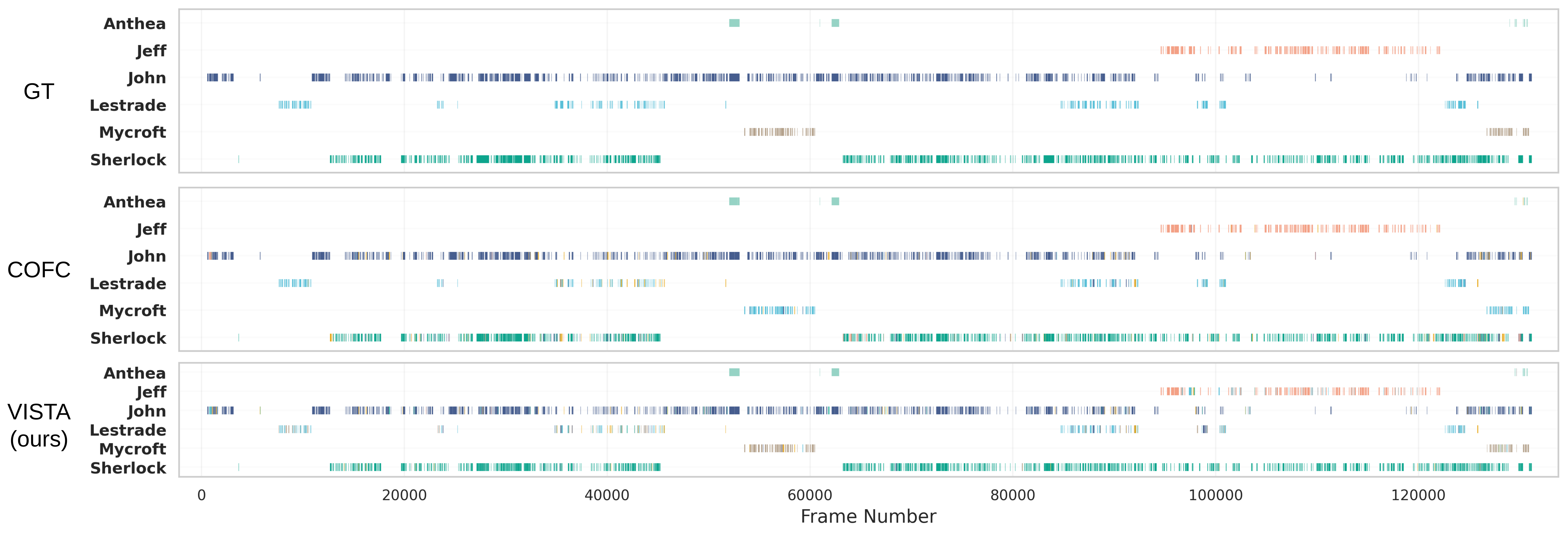}
    \caption{Timeline level qualitative comparison on \textit{Sherlock}. Top: ground truth; Middle: COFC; Bottom: VISTA. Each row represents a character, with colors indicating predicted clusters. Compared to COFC's cross identity confusion, VISTA maintains cleaner global identity organization and better separates visually similar characters (e.g., Lestrade and Mycroft).}
    \label{fig:qual_timeline}
\end{figure*}

\noindent

\subsection{Ablation Studies}
\label{sec:ablation}
We ablate key semantic components in Table~\ref{tab:ablation}. Overall, the results confirm that the semantic modules affect the final precision recall trade off, with the clearest gains appearing on \textit{TBBT} and \textit{Buffy}.
\begin{table}[h]
\centering
\caption{Throughput analysis of the VISTA pipeline}
\label{tab:throughput}
\resizebox{\columnwidth}{!}{
\begin{tabular}{lcccc}
\toprule
Dataset & Frames & Total Time (s) & End to End FPS & Clustering FPS \\
\midrule
\textit{TBBT} (s01e01) & 32,990 & 4,655 & 7.09 & 2122.45 \\
\textit{Buffy} (s05e01) & 62,620 & 7,978 & 7.85 & 2288.42 \\
\textit{Sherlock} (e01) & 132,150 & 16,393 & 8.06 & 1842.30 \\
\bottomrule
\end{tabular}}
\end{table}

\begin{table}[h]
\centering
\caption{Semantic arbitration invocation rate. CLIP/DeepFace attributes are extracted up front for every crop (Stage 1); this table measures how often the \emph{arbitration} step fires during association.}
\label{tab:vlm_intervention}
\resizebox{\columnwidth}{!}{
\begin{tabular}{lcccc}
\toprule
Dataset & \multicolumn{2}{c}{Frame level} & \multicolumn{2}{c}{Tracklet level} \\
\cmidrule(lr){2-3} \cmidrule(lr){4-5}
 & Arb. / Total & Rate & Arb. / Total & Rate \\
\midrule
\textit{TBBT}   & 29,389 / 319,646  & 9.19\%  & 3,685 / 20,737  & 17.77\% \\
\textit{Buffy}  & 64,222 / 474,210  & 13.54\% & 8,198 / 35,079  & 23.37\% \\
\textit{Sherlock} & 47,919 / 421,899 & 11.36\% & 5,797 / 30,448  & 19.04\% \\
\midrule
Overall & 141,530 / 1,215,755 & 11.64\% & 17,680 / 86,264 & 20.50\% \\
\bottomrule
\end{tabular}}
\end{table}

\paragraph{Semantic Components.}
As shown in Table~\ref{tab:ablation}, disabling Semantic Arbitration  or SSR   degrades performance (especially on \textit{TBBT}), confirming that geometric similarity alone struggles with appearance variations and that SSR is vital for stabilizing noisy predictions. Similarly, removing Three Zone Gating or Temporal Smoothing weakens the precision recall trade off, leading to brittle merges. Interestingly, on the volatile \textit{Sherlock} dataset, some ablated variants achieve higher individual metric scores. In particular, removing Temporal Smoothing raises WCP to $84.11\%$ (vs.\ $80.45\%$). We attribute this to Sherlock's frequent costume changes and disguises: temporal smoothing carries stale transient attribute profiles into subsequent decisions, occasionally penalising valid same identity merges. Conversely, removing SSR yields a higher NMI ($52.62\%$) and ID\_Acc ($71.43\%$) but at the cost of lower WCP ($78.36\%$), reflecting the fundamental purity completeness trade off. This dataset specific tension motivates future work on adaptive smoothing decay rates.

\paragraph{Effect of Tracklet Length.}
We ablate intra shot tracklet length averaged over all 15 episodes. Increasing the length from 12 to 48 improves ID\_Acc ($77.88\% \!\to\! 81.05\%$) and NMI ($77.39\% \!\to\! 78.72\%$), but extending to 96 slightly degrades both metrics due to accumulated appearance noise. WCP peaks at length 24 ($92.83\%$). We adopt $L_{\text{max}}=24$ as the unified default, balancing purity and the early decision requirement of online clustering. The full table is in the Supplementary Material.

\subsection{Qualitative Analysis}
\label{sec:qualitative}
Figure~\ref{fig:qual_timeline} ~and the crop level comparison in the Supplementary Material provide qualitative comparisons on \textit{Sherlock}, the most challenging dataset. These visualizations reveal that the main failure mode of classical online clustering is not uniform: some characters primarily suffer from cross identity confusion, while others are more affected by trajectory fragmentation.

Figure~\ref{fig:qual_timeline} shows that COFC's identity errors extend beyond local boundary noise: predicted clusters intrude into other characters' timelines for extended intervals. A representative case is COFC's Lestrade timeline, which contains repeated intrusions from a Mycroft dominant cluster, indicating substantial cross identity confusion rather than mere short term fragmentation. VISTA preserves cleaner separation between these visually similar but semantically distinct characters, suggesting that semantic triage prevents overly aggressive merging.

~The crop level comparison (Supplementary Material) complements this: under challenging pose, illumination, and blur conditions, COFC more frequently assigns samples to competing clusters, whereas VISTA keeps a larger portion within identity consistent clusters. Importantly, VISTA's remaining errors on \textit{Sherlock} stem from overly conservative semantic decisions that occasionally fragment the same identity into multiple segments, rather than cross identity confusion. This is precisely the mechanism behind the ID\_Acc decrease relative to COFC on \textit{Sherlock} (Table~\ref{tab:main_results}): purity is defended at the cost of identity continuity. The qualitative gain is thus better understood as a shift in error profile, from large scale identity mixing toward more structured, localized fragmentation, consistent with the quantitative trends in Table~\ref{tab:ablation}. This also explains why removing Temporal Smoothing can raise WCP on \textit{Sherlock}: without stale attribute profiles, fewer valid merges are rejected, at the cost of reduced cross identity discrimination.

\noindent

\subsection{Computational Efficiency and Throughput}
\label{sec:efficiency}
A critical requirement for online clustering is computational efficiency. To demonstrate that VISTA's semantic triage and spatio temporal arbitration do not introduce prohibitive overhead, we profile the throughput of our pipeline. As shown in Table~\ref{tab:throughput}, the performance exhibits a clear dichotomy between the heavy feature extraction front end and the  efficient clustering back end.

The end to end throughput (Total FPS) stabilizes at approximately 7 to 8 frames per second. Profiling reveals that over 99\% of this time is consumed by Stage 1 (Feature Extraction), which involves decoding high resolution video frames and executing dense face detection  alongside multi-modal feature encoding (FaceNet, CLIP, and DeepFace attribute extraction).

In contrast, VISTA's online clustering module (Stages 2 and 3) operates at 1800 to 2300 FPS. This highlights a major advantage of our design: by abstracting raw VLM outputs into compact SSR and employing dynamic GMM thresholding, VISTA restricts semantic arbitration to ambiguous boundary cases. Consequently, the proposed vision language integration achieves robust performance gains while keeping the online clustering phase itself  efficient; in our implementation, the dominant runtime bottleneck remains face detection and feature extraction and not the clustering.

The Spatio Temporal Semantic Triage acts as an effective decision filter (Table~\ref{tab:vlm_intervention}). Across 1.2M+ frames, semantic arbitration is invoked for only $20.50\%$ of tracklets (covering $11.64\%$ of total frames). \textit{TBBT} requires the least intervention ($9.19\%$ of frames) due to cleaner visual conditions, while \textit{Buffy}'s darker scenes trigger more ambiguity ($13.54\%$).  Since CLIP/DeepFace attributes are extracted up front for every crop in Stage 1 (always included in the end to end cost), Table~\ref{tab:vlm_intervention} measures how often arbitration fires during association, not how many VLM forward passes are avoided; triage keeps the \emph{clustering} stage lightweight by restricting arbitration to ambiguous boundary cases.

\subsection{Analysis of Demographic Disparities}
\label{sec:bias_analysis}
\begin{figure}[htbp]
    \centering
    \begin{subfigure}[b]{0.49\linewidth}
        \centering
        \includegraphics[width=\linewidth]{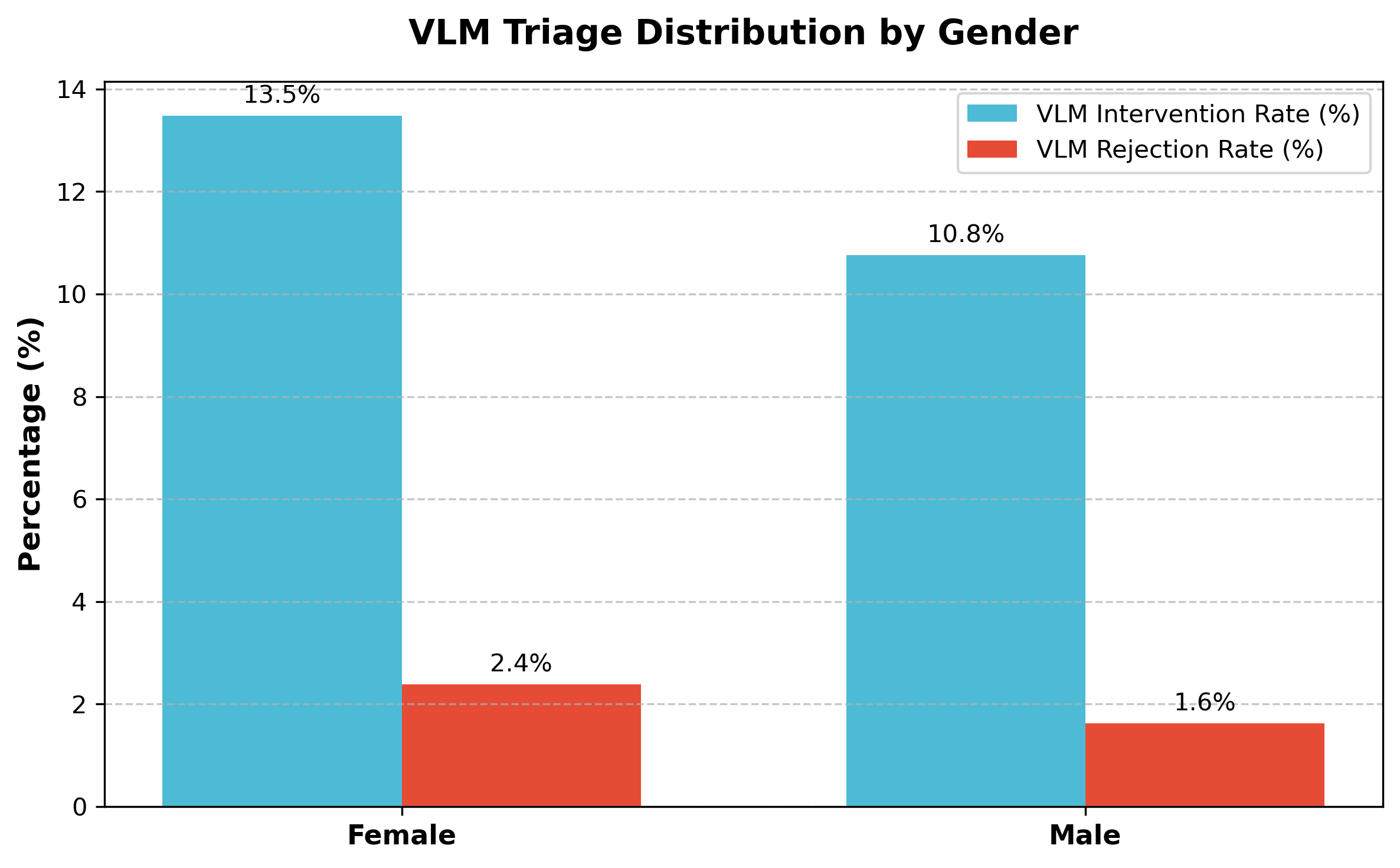}
        \caption{By Gender}
        \label{fig:bias_gender}
    \end{subfigure}
    \hfill
    \begin{subfigure}[b]{0.49\linewidth}
        \centering
        \includegraphics[width=\linewidth]{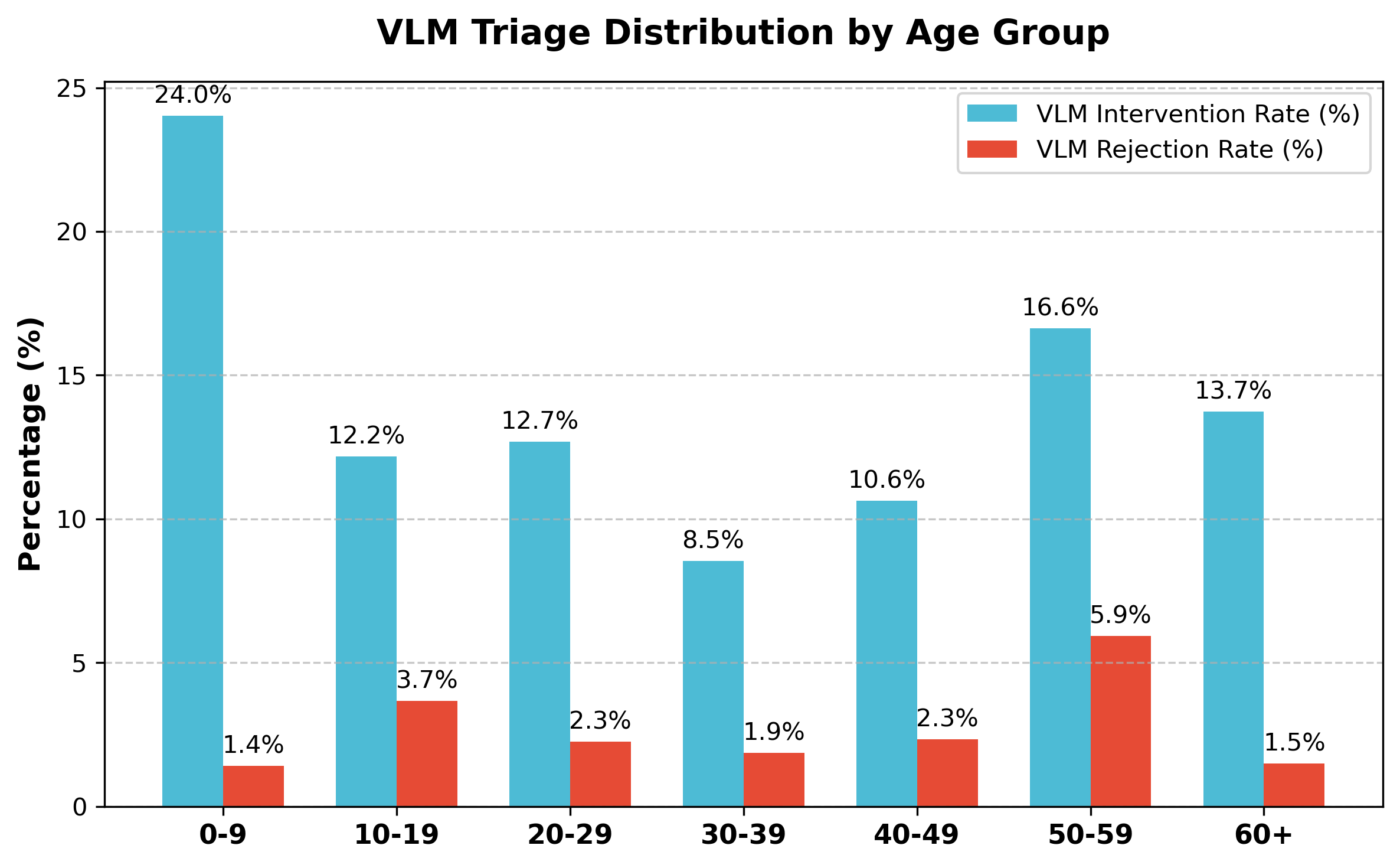}
        \caption{By Age Group}
        \label{fig:bias_age}
    \end{subfigure}

    \caption{Distribution of VLM intervention and rejection rates across different demographic groups.}
    \label{fig:demographic_bias}
\end{figure}

Although VISTA reduces global identity confusion, VLM arbitration is not invoked uniformly across predicted demographic groups. We therefore report both triage rate diagnostics (Figure~\ref{fig:demographic_bias}) and subgroup clustering-quality gaps in the Supplementary Material. Female characters show higher intervention and ambiguity-zone rejection rates than male characters (13.48\% vs.\ 10.76\%, and 2.38\% vs.\ 1.62\%), suggesting that more disambiguation burden is offloaded to semantic arbitration in unconstrained settings. This observation is consistent with known gender accuracy disparities in commercial vision models~\cite{buolamwini2018gender}, although our analysis is diagnostic rather than causal.

\textbf{Age Disparities:} Figure~\ref{fig:bias_age} reveals distinct effects at age extremes. Children (0--9 years) trigger the highest VLM intervention rate (24.03\%), while older adults (50--59 years) face the highest rejection rate (5.93\%). Conversely, the core 30--39 demographic, which is common in face model training distributions~\cite{karkkainen2021fairface}, requires the least intervention (8.55\%). These trends suggest that skewed training data and difficult capture conditions can jointly affect representation robustness for age extremes~\cite{suresh2021framework}.

\textbf{Subgroup Quality and Safeguards:} Beyond triage rates, the Supplementary Material reports WCP/NMI/ID\_Acc gaps on the IoU matched subset using the same VLM predicted gender/age groups. Because VPCD lacks ground truth demographics, these groups do not constitute a full fairness audit; nevertheless, the gap signs vary across datasets rather than showing one directional degradation. As an ethical safeguard, VISTA can disable gender/age cues and fall back to geometry plus transient attributes.

\textbf{Discussion:} These preliminary findings show clear algorithmic disparities, but demographic variables in videos often entangle with environmental factors. Screen industry tropes, visual ageism~\cite{crawford2020excavating, ozimek2020equality, calasanti2006ageism}, and historical underrepresentation~\cite{smith2020inequality, bamman2024measuring} often dictate lighting, camera angles, and action sequences for specific groups. For instance, low illumination scenes common in action genres can disproportionately degrade geometric feature extraction for darker skin tones, compounding existing gender and age ambiguities. Thus, attributing these disparities solely to intrinsic model bias requires further disentanglement. Future work will systematically analyze these issues by cross referencing scene level factors (illumination, pose, focal presence) with face model training distributions, while extending to the intersectional impacts of digital racialization~\cite{amaro2022black, sharma2023understanding}. Addressing these entangled biases is essential for deploying equitable face clustering systems in real world, unconstrained video streaming applications.

\section{Conclusion}

We presented VISTA, a vision language semantic triage architecture for fully online face clustering in long videos. By combining Symbolic Semantic Representation, dynamic GMM thresholding, and spatio temporal decoupled arbitration, VISTA mitigates over segmentation and improves the purity--continuity trade off on \textit{TBBT}, \textit{Buffy}, and \textit{Sherlock}. Under identical detector, embeddings, and evaluation protocol, VISTA consistently outperforms the strongest matched online unsupervised baseline while preserving streaming compatible efficiency.
\textbf{Limitations and Future Work:} VISTA still relies on fixed symbolic vocabularies, a bimodal GMM prior, and approximate confidence attenuation; cross domain validation beyond scripted TV remains open. On \textit{Sherlock}, VISTA trades identity continuity for purity (ID\_Acc $-3.17$\,pp vs.\ COFC), motivating adaptive smoothing, calibrated VLM uncertainty, and confounder aware fairness analysis. 
\begin{acks}
This work was funded by the Leverhulme Trust (RPG-2023-091).
\end{acks}

\bibliographystyle{ACM-Reference-Format}
\balance %
\bibliography{sample-base}

\end{document}


\title{Supplementary Material:\\ VISTA: Vision-Language Integrated Semantic Triage Architecture for Online Face Clustering in Long Videos}

\author{Jiashu Liao}
\orcid{0000-0002-1058-1641}
\affiliation{%
  \institution{University of Glasgow}
  \city{Glasgow}
  \country{United Kingdom}}
\email{jiashu.liao@glasgow.ac.uk}

\author{Joanne Garde-Hansen}
\orcid{0000-0003-2462-3790}
\affiliation{%
  \institution{University of Leeds}
  \city{Leeds}
  \country{United Kingdom}}
\email{j.garde-hansen@leeds.ac.uk}

\author{Sanjay Sharma}
\orcid{0000-0003-2898-9925}
\affiliation{%
  \institution{The University of Warwick}
  \city{Coventry}
  \country{United Kingdom}}
\email{sanjay.sharma@warwick.ac.uk}

\author{Tanaya Guha}
\orcid{0000-0003-2167-4891}
\affiliation{%
  \institution{University of Glasgow}
  \city{Glasgow}
  \country{United Kingdom}}
\email{tanaya.guha@glasgow.ac.uk}

\renewcommand{\shortauthors}{Jiashu Liao, Joanne Garde-Hansen, Sanjay Sharma, and Tanaya Guha}

\maketitle

\section{Supplementary Video Demonstrations}
We provide two supplementary video demonstrations, \texttt{demo\allowbreak\_mycroft\allowbreak.mp4} and \texttt{demo\allowbreak\_lestrade\allowbreak.mp4}, to illustrate the temporal stability of our framework across different levels of scene complexity. Both videos have been compressed to meet the file size limit of the submission system, but still clearly show a side-by-side comparison between the COFC baseline and VISTA in \textit{Sherlock} (Season 1, Episode 1).

\paragraph{Video Layout \& Evaluation Metric.}
Both videos are rendered in a condensed highlight format, with the left panel showing COFC and the right panel showing VISTA. We color the bounding boxes dynamically based on a purity-based metric:
\begin{itemize}
    \item \textbf{Cyan Box (Correct):} The predicted cluster's majority identity matches the ground-truth target character, and the cluster purity exceeds our threshold ($\ge 60\%$).
    \item \textbf{Red Box (Wrong):} The cluster is either misclassified (wrong majority identity) or heavily contaminated (purity $< 60\%$). A contaminated cluster is treated as a failure even if it contains the target character.
\end{itemize}

\paragraph{Demo 1: Resolving Identity Over-merging.}
This video (\texttt{demo\allowbreak\_mycroft\allowbreak.mp4}) focuses on \textit{Mycroft} in relatively stable conversational scenes.
\begin{itemize}
    \item \textbf{Baseline:} COFC over-merges \textit{Mycroft} and \textit{Lestrade} into a single large cluster. Although the assigned ID appears temporally continuous, it is incorrect, producing a persistent red box throughout.
    \item \textbf{VISTA:} By leveraging vision-language semantic arbitration, VISTA distinguishes the two characters and isolates \textit{Mycroft} into a separate high-purity cluster, as indicated by the consistent cyan box.
\end{itemize}

\paragraph{Demo 2: Purity vs.\ Continuity in Dynamic Scenes.}
This video (\texttt{demo\allowbreak\_lestrade\allowbreak.mp4}) tracks \textit{Lestrade} through scenes with frequent occlusions, motion blur, and illumination changes.
\begin{itemize}
    \item \textbf{Baseline:} COFC merges \textit{Lestrade}, \textit{Mycroft}, and several background extras into a single contaminated cluster (purity ${\sim}55\%$), which fails our metric and remains red throughout.
    \item \textbf{VISTA:} To preserve cluster purity under these conditions, VISTA's dynamic triage deliberately over-splits the track into a few separate but pure clusters (e.g., C22 and C26, both with purity ${>}60\%$). This trade-off ensures that downstream offline aggregation receives clean tracklets rather than contaminated ones. The cyan boxes confirm that each resulting cluster passes our threshold.
\end{itemize}

\section{VISTA Algorithm Pseudocode}
To provide a clearer understanding of the online clustering process, we detail the step by step execution of VISTA in Algorithm~\ref{alg:vista}. The algorithm processes the video stream in a single pass. For each camera shot, it extracts dense visual features and zero shot VLM attributes. Short term temporal tracking is performed to construct reliable local tracklets. Subsequently, the inter shot association is guided by our dynamic Gaussian Mixture Model (GMM) thresholding and spatio temporal decoupled semantic arbitration. This pipeline ensures that low confidence semantic noise is filtered out while valuable temporal constraints and geometric similarities are preserved. We emphasise that geometric embeddings and CLIP/DeepFace attributes are extracted up front for every face crop at shot ingestion (line 3) and voted into tracklet level SSR profiles; the three zone triage controls only whether semantic \emph{arbitration} is invoked during association, not whether VLM inference is performed. The final acceptance gate for ambiguity zone pairs is $\tau_{\text{merge}}=0.325$ (see Section 3.3 of the main paper).

\begin{algorithm}[h]
\caption{VISTA Online Clustering Pipeline}
\label{alg:vista}
\begin{algorithmic}[1]
\REQUIRE Video stream $\mathcal{V}$, initial thresholds $\tau_{\text{high}}^{(0)}, \tau_{\text{low}}^{(0)}, \tau_{\text{merge}}{}=0.325$
\ENSURE Global identity clusters $\mathcal{C}$
\STATE Initialize $\mathcal{C} \leftarrow \emptyset$, similarity queue $\mathcal{Q} \leftarrow \emptyset$
\FOR{each shot $S$ in $\mathcal{V}$}
 \STATE Extract faces, compute geometric embeddings $v$ and VLM attributes $\mathcal{A}$ (for every face crop)
 \STATE Form intra shot tracks $\{T_1, \ldots, T_m\}$; build temporal constraint $Q$-matrix
 \FOR{each track $T_i$}
   \STATE Compute SSR profile via temporal majority vote
   \STATE Find best matching cluster: $C^* = \arg\max_{C_j} \mathrm{sim}(v_{T_i}, v_{C_j})$ s.t.\ $Q_{T_i, C_j} = 1$
   \STATE $s \leftarrow \mathrm{sim}(v_{T_i}, v_{C^*})$
   \IF{$s > \tau_{\text{high}}$}
     \STATE Apply bonus only semantic adjustment; merge $T_i$ into $C^*$
   \ELSIF{$s \ge \tau_{\text{low}}$}
     \STATE Compute adjustment $\mathrm{adj}$ via Spatio Temporal Decoupled Arbitration
     \IF{$s + \mathrm{adj} \ge \tau_{\text{merge}}$}
       \STATE Merge $T_i$ into $C^*$ (Resolved ambiguity)
     \ELSE
       \STATE Create new cluster $C_{\text{new}} \leftarrow \{T_i\}$; $\mathcal{C} \leftarrow \mathcal{C} \cup \{C_{\text{new}}\}$
     \ENDIF
   \ELSE
     \IF{gender + age strongly agree (persistent trait rescue)}
       \STATE Merge $T_i$ into $C^*$ (Rescued from reject zone)
     \ELSE
       \STATE Create new cluster $C_{\text{new}} \leftarrow \{T_i\}$; $\mathcal{C} \leftarrow \mathcal{C} \cup \{C_{\text{new}}\}$
     \ENDIF
   \ENDIF
   \STATE Push $s$ to $\mathcal{Q}$; update GMM thresholds ($\tau_{\text{high}}, \tau_{\text{low}}$) via EMA ($\eta_{\text{gmm}}=0.3$)
 \ENDFOR
\ENDFOR
\RETURN $\mathcal{C}$
\end{algorithmic}
\end{algorithm}

\section{Age Symbol Configurations}
As discussed in Section 3.3 of the main paper, applying raw VLM textual predictions directly introduces severe frame level noise. To mitigate this, VISTA maps raw age strings to a discrete vocabulary. We tested three different age grouping strategies, detailed in Table~\ref{tab:age_configs}.

The \texttt{fine\_grained} strategy preserves decade level precision for adults, which can occasionally differentiate close identities but is highly susceptible to VLM oscillation (e.g., predicting ``20-29'' in one frame and ``30-39'' in the next). The \texttt{broad\_adult} strategy collapses a massive 40-year span into a single token, which guarantees stability but loses discriminative power. The coarse\_grouping strategy provides the most robust compromise
for the unified setting by reducing age boundary oscillation without
collapsing the entire adult range into a single token.

\begin{table}[h]
\centering
\caption{Age symbol configurations. Raw DeepFace age predictions are mapped to discrete tokens via one of three schemes. \texttt{fine\_grained} preserves decade level distinctions for adults; \texttt{coarse\_grouping} merges the 30s and 40s into a single \texttt{<age\_adult>} token; \texttt{broad\_adult} further collapses the full 20-59 range.}
\label{tab:age_configs}
\resizebox{\columnwidth}{!}{
\begin{tabular}{lccccc}
\toprule
Raw age range & \texttt{fine\_grained} & \texttt{coarse\_grouping} & \texttt{broad\_adult} \\
\midrule
0--9   & \texttt{<age\_child>}  & \texttt{<age\_child>}  & \texttt{<age\_child>}  \\
10--19 & \texttt{<age\_teen>}   & \texttt{<age\_teen>}   & \texttt{<age\_teen>}   \\
20--29 & \texttt{<age\_20s>}    & \texttt{<age\_young\_adult>} & \texttt{<age\_adult>} \\
30--39 & \texttt{<age\_30s>}    & \texttt{<age\_adult>}  & \texttt{<age\_adult>} \\
40--49 & \texttt{<age\_40s>}    & \texttt{<age\_adult>}  & \texttt{<age\_adult>} \\
50--59 & \texttt{<age\_50s>}    & \texttt{<age\_senior>} & \texttt{<age\_adult>} \\
60+    & \texttt{<age\_senior>} & \texttt{<age\_senior>} & \texttt{<age\_senior>} \\
\midrule
Vocabulary size & 7 & 5 & 4 \\
\bottomrule
\end{tabular}}
\end{table}

\section{CLIP Prompt Templates and Symbol Registry}

VISTA uses CLIP ViT-B/32 for zero shot attribute classification. For each detected face crop, logits are computed against all prompt candidates simultaneously in a single forward pass. Table~\ref{tab:prompts} lists the complete prompt templates and label vocabularies for each attribute dimension. DeepFace is used separately for age estimation (not via CLIP prompts) and its raw integer predictions are mapped through the age symbol configurations in Table~\ref{tab:age_configs}.

\begin{table}[h]
\centering
\caption{CLIP zero shot classification prompts and label vocabularies.}
\label{tab:prompts}
\resizebox{\columnwidth}{!}{
\begin{tabular}{lll}
\toprule
Attribute & Prompt Template & Labels \\
\midrule
Gender & ``A face of a \{label\}'' & Male, Female \\
\midrule
\multirow{2}{*}{Clothing} & \multirow{2}{*}{``A photo of a person wearing a \{label\}''} & T-shirt, Jacket, Shirt, Sweater, \\
 & & Dress, Coat, Vest \\
\midrule
\multirow{2}{*}{Hair} & \multirow{2}{*}{``A close-up photo of a person with \{label\}''} & Curly Hair, Straight Hair, Bald, \\
 & & Short Hair, Long Hair \\
\bottomrule
\end{tabular}}
\end{table}

Table~\ref{tab:symbol_registry} provides the complete Symbol Registry mappings that convert raw CLIP predictions into stable symbolic tokens (SSR). Gender and hair labels are mapped one to one, while clothing labels undergo a many to one compression (e.g., both ``T-shirt'' and ``Shirt'' map to distinct tokens, but the vocabulary is reduced from open ended text to a fixed set of 7 symbols). Age mappings are handled separately via the configurations in Table~\ref{tab:age_configs}.

\begin{table}[h]
\centering
\caption{Symbol Registry: raw label to symbolic token mappings.}
\label{tab:symbol_registry}
\resizebox{\columnwidth}{!}{
\begin{tabular}{lll}
\toprule
Attribute & Raw Label & Symbolic Token \\
\midrule
\multirow{2}{*}{Gender} & Male & \texttt{<gender\_male>} \\
 & Female & \texttt{<gender\_female>} \\
\midrule
\multirow{7}{*}{Clothing} & T-shirt & \texttt{<clothing\_tshirt>} \\
 & Jacket & \texttt{<clothing\_jacket>} \\
 & Shirt & \texttt{<clothing\_shirt>} \\
 & Sweater & \texttt{<clothing\_sweater>} \\
 & Dress & \texttt{<clothing\_dress>} \\
 & Coat & \texttt{<clothing\_coat>} \\
 & Vest & \texttt{<clothing\_vest>} \\
\midrule
\multirow{5}{*}{Hair} & Curly Hair & \texttt{<hair\_curly>} \\
 & Straight Hair & \texttt{<hair\_straight>} \\
 & Bald & \texttt{<hair\_bald>} \\
 & Short Hair & \texttt{<hair\_short>} \\
 & Long Hair & \texttt{<hair\_long>} \\
\bottomrule
\end{tabular}}
\end{table}

For age estimation, VISTA uses DeepFace~\cite{Serengil2024} with \linebreak
\texttt{detector\_backend='skip'} (since face crops are
already provided by RetinaFace) and \texttt{actions=['age']}. The raw integer age prediction is binned into decade ranges (e.g., 25 $\to$ ``20-29'') and then mapped through the Symbol Registry age configuration (Table~\ref{tab:age_configs}). In the unified evaluation, we use the \texttt{coarse\_grouping} preset, yielding a 5-symbol age vocabulary.

\section{Dataset Specific Parameter Effects}
While the main paper reports VISTA's performance using a unified parameter setting for fair benchmarking, we present a marginal sensitivity analysis here to demonstrate the framework's robustness across different visual environments. Tables~\ref{tab:param_effect_tbbt}, \ref{tab:param_effect_buffy}, and \ref{tab:param_effect_sherlock} display the marginal effects of individual hyperparameters on \textit{TBBT}, \textit{Buffy}, and \textit{Sherlock}, respectively.

Overall, the standard deviation of metrics across these reasonable hyperparameter ranges remains small. We consistently observe that a higher cluster representation smoothing momentum ($\texttt{ema\_rep} = 0.95$; named \texttt{ema} in an earlier version, distinct from the GMM threshold smoothing coefficient $\eta_{\text{gmm}}$) is beneficial for preventing abrupt threshold oscillation in online scenarios. Furthermore, challenging datasets with darker scenes (e.g., \textit{Buffy}) favor a larger GMM sliding window ($\texttt{queue} = 300$) to gather enough diverse samples for stable threshold estimation, whereas visually stable datasets (e.g., \textit{TBBT}) can adapt faster with a smaller queue. Note that the $\tau_{\text{high}}$ spread in these tables reflects simultaneous co-variation of $\texttt{ema\_rep}$ and queue size; these marginal tables are not a one factor test of $\tau_{\text{high}}$ alone. A dedicated $3\times3$ warm up threshold grid with all other settings fixed is reported in Section~\ref{sec:threshold_grid}.

\begin{table}[ht]
\centering
\caption{\textit{TBBT} parameter effects under dataset specific marginal analysis. Values indicate average performance when the specific parameter is fixed while others vary.}
\label{tab:param_effect_tbbt}
\resizebox{\columnwidth}{!}{
\begin{tabular}{lccc}
\toprule
Parameter & Value & WCP / NMI / ID\_Acc & Bal. Avg. \\
age & broad\_adult & 96.66 / \textbf{87.57} / \textbf{89.40} & \textbf{91.21} \\
age & coarse\_grouping & \textbf{96.70} / 87.15 / 88.58 & 90.81 \\
age & fine\_grained & 96.60 / 87.01 / 88.68 & 90.76 \\
\midrule
ema\_rep & 0.95 & \textbf{97.47} / \textbf{89.43} / \textbf{91.32} & \textbf{92.74} \\
ema\_rep & 0.75 & 95.84 / 85.05 / 86.46 & 89.11 \\
\midrule
queue & 300.0 & \textbf{96.97} / \textbf{87.65} / \textbf{89.17} & \textbf{91.26} \\
queue & 150.0 & 96.34 / 86.83 / 88.60 & 90.59 \\
\midrule
vishigh & 0.75 & \textbf{96.65} / \textbf{87.24} / \textbf{88.89} & \textbf{90.93} \\
vishigh & 0.825 & \textbf{96.65} / \textbf{87.24} / \textbf{88.89} & \textbf{90.93} \\
vishigh & 0.9 & \textbf{96.65} / \textbf{87.24} / \textbf{88.89} & \textbf{90.93} \\
\midrule
vislow & 0.275 & 96.64 / \textbf{87.30} / \textbf{88.97} & \textbf{90.97} \\
vislow & 0.2 & 96.65 / 87.21 / 88.88 & 90.91 \\
vislow & 0.35 & \textbf{96.67} / 87.20 / 88.82 & 90.90 \\
\bottomrule
\end{tabular}}
\end{table}

\begin{table}[ht]
\centering
\caption{\textit{Buffy} parameter effects under dataset specific marginal analysis.}
\label{tab:param_effect_buffy}
\resizebox{\columnwidth}{!}{
\begin{tabular}{lccc}
\toprule
Parameter & Value & WCP / NMI / ID\_Acc & Bal. Avg. \\
age & coarse\_grouping & \textbf{84.85} / \textbf{75.24} / \textbf{76.98} & \textbf{79.02} \\
age & broad\_adult & 84.18 / 74.50 / 76.02 & 78.23 \\
age & fine\_grained & 81.62 / 72.17 / 73.49 & 75.76 \\
\midrule
ema\_rep & 0.95 & \textbf{85.18} / \textbf{77.22} / \textbf{79.68} & \textbf{80.69} \\
ema\_rep & 0.75 & 81.87 / 70.67 / 71.24 & 74.59 \\
\midrule
queue & 150.0 & \textbf{84.70} / \textbf{75.01} / \textbf{76.31} & \textbf{78.67} \\
queue & 300.0 & 82.39 / 72.93 / 74.67 & 76.66 \\
\midrule
vishigh & 0.825 & \textbf{84.67} / \textbf{75.04} / \textbf{76.57} & \textbf{78.76} \\
vishigh & 0.9 & \textbf{84.67} / \textbf{75.04} / \textbf{76.57} & \textbf{78.76} \\
vishigh & 0.75 & 81.31 / 71.85 / 73.36 & 75.51 \\
\midrule
vislow & 0.35 & 84.60 / \textbf{75.08} / \textbf{76.72} & \textbf{78.80} \\
vislow & 0.2 & \textbf{84.86} / 75.05 / 76.48 & \textbf{78.80} \\
vislow & 0.275 & 81.20 / 71.79 / 73.30 & 75.43 \\
\bottomrule
\end{tabular}}
\end{table}

\begin{table}[ht]
\centering
\caption{\textit{Sherlock} parameter effects under dataset specific marginal analysis.}
\label{tab:param_effect_sherlock}
\resizebox{\columnwidth}{!}{
\begin{tabular}{lccc}
\toprule
Parameter & Value & WCP / NMI / ID\_Acc & Bal. Avg. \\
age & broad\_adult & 80.16 / \textbf{46.69} / \textbf{58.51} & \textbf{61.79} \\
age & fine\_grained & 80.23 / 46.68 / 58.08 & 61.66 \\
age & coarse\_grouping & \textbf{80.38} / 46.67 / 55.58 & 60.87 \\
\midrule
ema\_rep & 0.95 & \textbf{81.88} / \textbf{49.36} / \textbf{61.54} & \textbf{64.26} \\
ema\_rep & 0.75 & 78.63 / 44.00 / 53.24 & 58.62 \\
\midrule
queue & 150.0 & 80.04 / \textbf{46.72} / \textbf{58.56} & \textbf{61.77} \\
queue & 300.0 & \textbf{80.47} / 46.64 / 56.22 & 61.11 \\
\midrule
vishigh & 0.75 & \textbf{80.25} / \textbf{46.68} / \textbf{57.39} & \textbf{61.44} \\
vishigh & 0.825 & \textbf{80.25} / \textbf{46.68} / \textbf{57.39} & \textbf{61.44} \\
vishigh & 0.9 & \textbf{80.25} / \textbf{46.68} / \textbf{57.39} & \textbf{61.44} \\
\midrule
vislow & 0.2 & \textbf{80.41} / \textbf{46.82} / \textbf{57.69} & \textbf{61.64} \\
vislow & 0.275 & 80.27 / 46.74 / 57.57 & 61.53 \\
vislow & 0.35 & 80.08 / 46.48 / 56.92 & 61.16 \\
\bottomrule
\end{tabular}}
\end{table}

\section{Semantic Arbitration Invocation Rate Visualization}

\begin{figure*}[ht]
\centering
\includegraphics[width=0.24\textwidth]{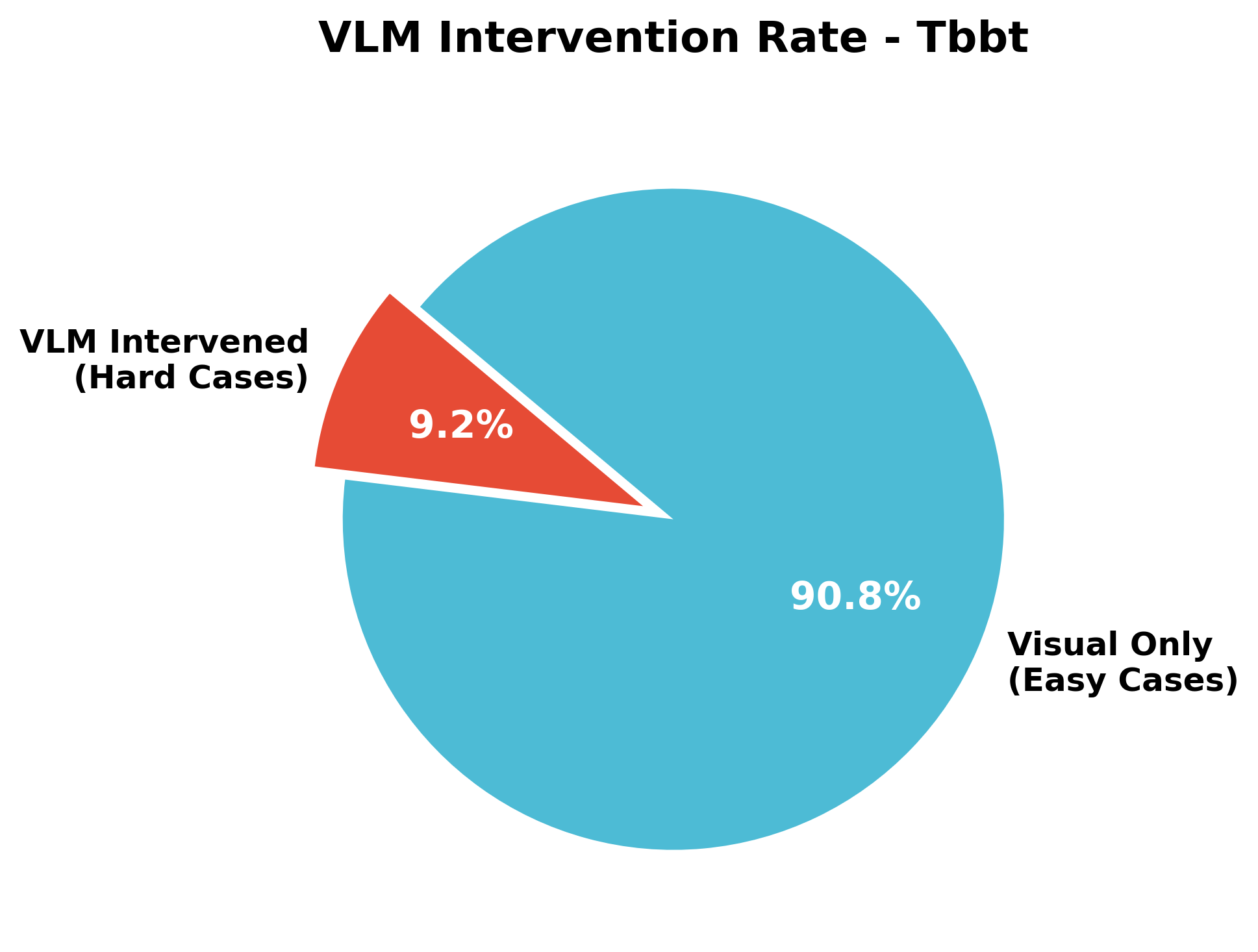}
\includegraphics[width=0.24\textwidth]{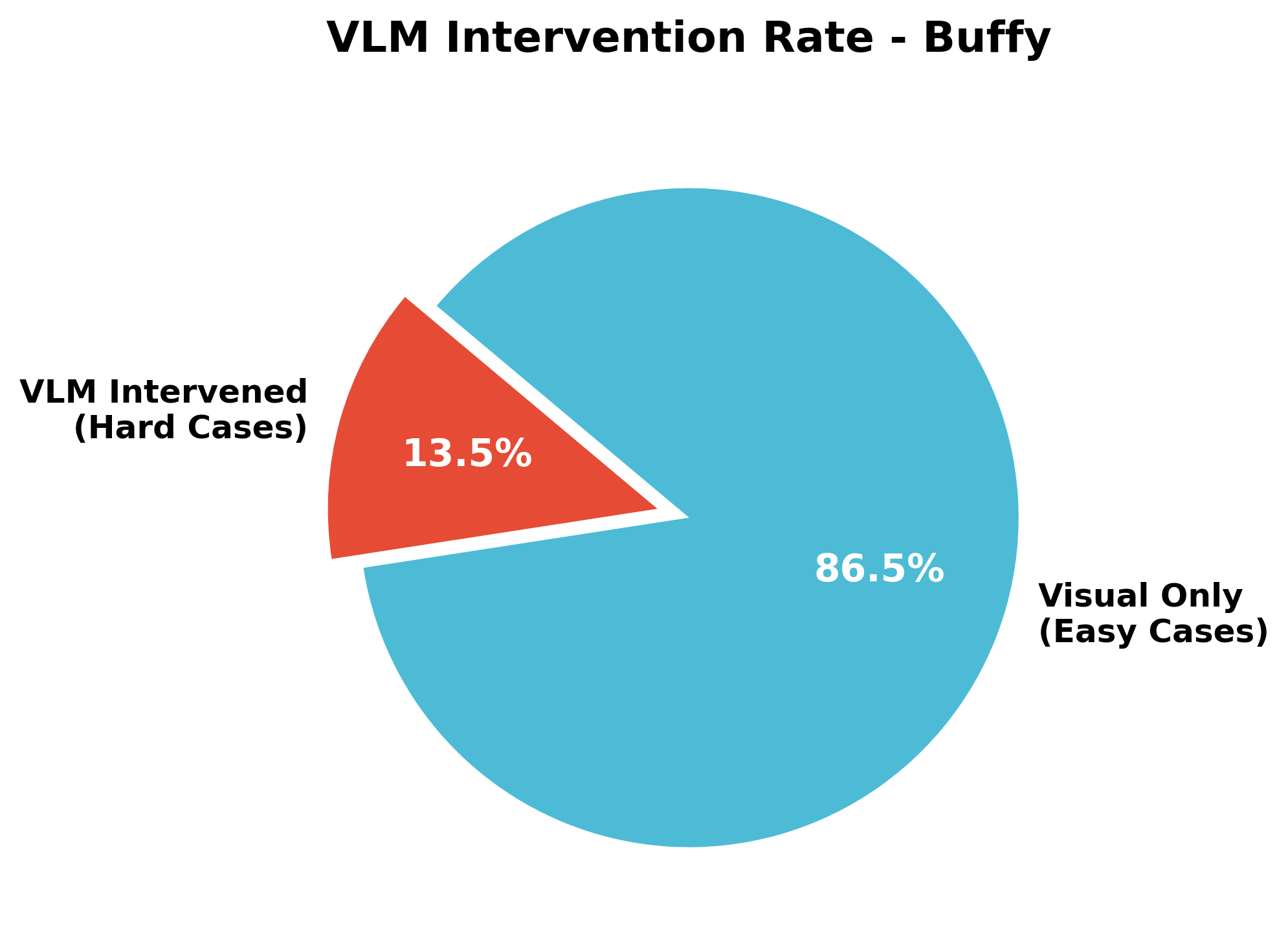}
\includegraphics[width=0.24\textwidth]{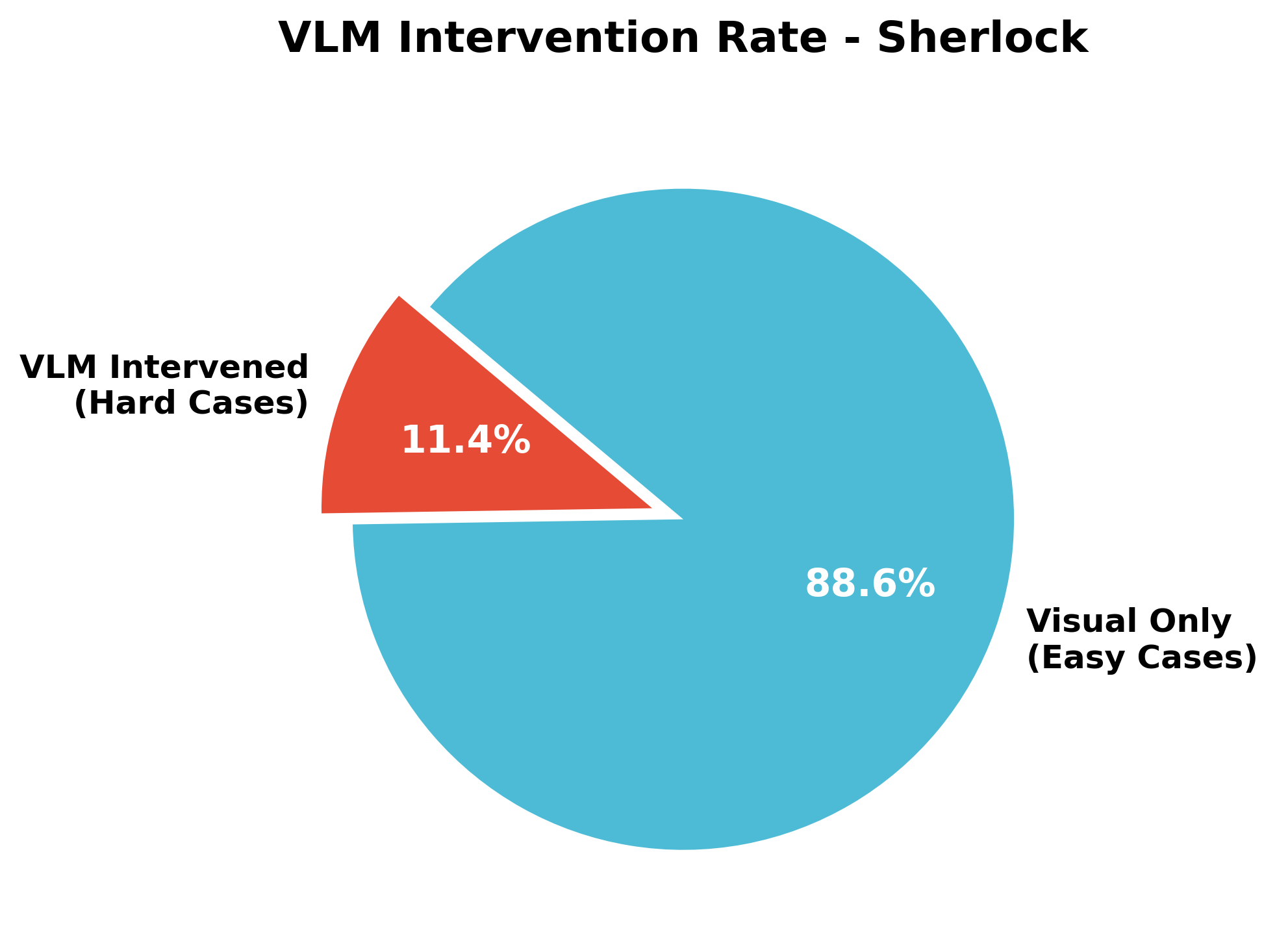}
\includegraphics[width=0.24\textwidth]{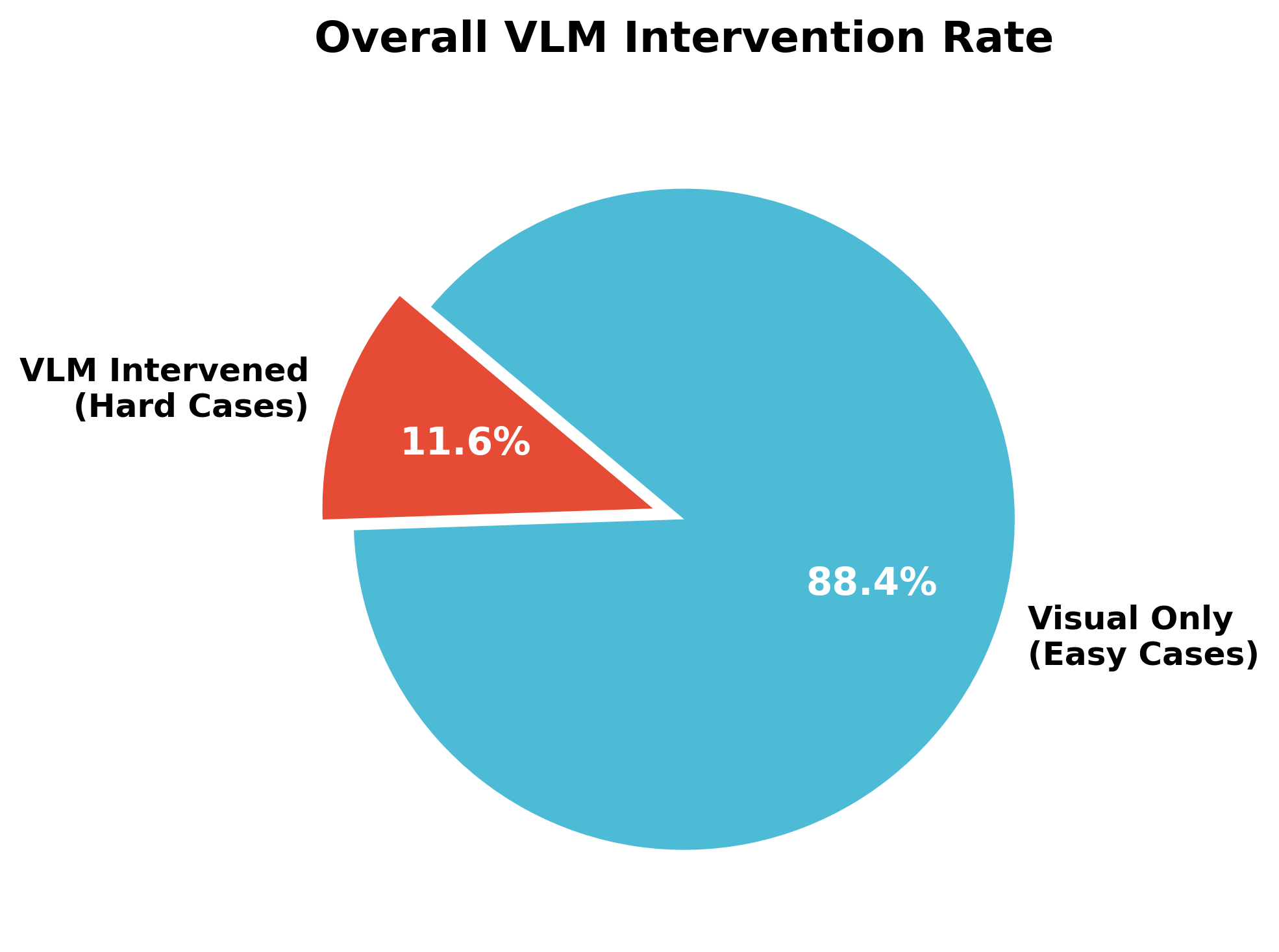}
\caption{Frame level semantic arbitration invocation rate per dataset and overall. The dynamic triage mechanism restricts arbitration to a small fraction of ambiguous frames, preserving near real time processing speeds.}
\label{fig:vlm_pie}
\end{figure*}

Figure~\ref{fig:vlm_pie} visually reinforces the high efficiency of VISTA, complementing the raw statistics reported in the main paper. Across all three datasets, the triage mechanism successfully limits semantic arbitration by routing the vast majority of frames ($86.5\%$-$90.8\%$) through the fast, visual only pipeline. As clarified in the main paper (Section 4.6), CLIP/DeepFace attributes are extracted up front for every face crop; these rates measure how often the arbitration step fires during association, not how many VLM forward passes are avoided. Notably, \textit{TBBT} requires the least arbitration ($9.2\%$), consistent with its well lit studio conditions. Conversely, \textit{Buffy} triggers the highest rate ($13.5\%$) due to frequent low illumination scenes and fast motion that push more track cluster pairs into the ambiguity zone. \textit{Sherlock} falls between the two ($11.4\%$) despite being the hardest dataset overall, suggesting that its difficulty stems more from extreme long range appearance variation (e.g., character disguises) than from persistent local visual ambiguity.

\section{Semantic Arbitration Weight Sensitivity}

To systematically assess the robustness of the semantic arbitration function introduced in the main paper, we perform a comprehensive grid search over 243 configurations covering all five semantic weight dimensions:
\begin{align*}
    \lambda_{\text{ga}} &\in \{0.04, 0.08, 0.12\}, & \lambda_{\text{cloth}} &\in \{0.02, 0.05, 0.08\}, \\
    \lambda_{\text{gm}} &\in \{0.08, 0.12, 0.16\}, & \lambda_{\text{am}} &\in \{0.01, 0.03, 0.05\}, \\
    \lambda_{\text{hc}} &\in \{0.02, 0.04, 0.06\}.
\end{align*}

\paragraph{Overall stability.}
Table~\ref{tab:lambda_variance} reports the minimum, maximum, and standard deviation of WCP and NMI across all 243 configurations for each dataset. The remarkably low standard deviations (e.g., only 0.31 for WCP on \textit{TBBT}) confirm that VISTA's clustering gains are not the product of fragile weight tuning.

\begin{table}[ht]
\centering
\caption{Performance variance across 243 semantic weight ($\lambda$) configurations.}
\label{tab:lambda_variance}
{
\begin{tabular}{lcccc}
\toprule
Dataset & Metric & Min & Max & Std \\
\midrule
\multirow{2}{*}{\textit{TBBT}} & WCP & 96.08 & 97.59 & 0.31 \\
 & NMI & 90.20 & 91.79 & 0.27 \\
\midrule
\multirow{2}{*}{\textit{Buffy}} & WCP & 84.71 & 90.34 & 1.09 \\
 & NMI & 76.86 & 80.96 & 0.74 \\
\midrule
\multirow{2}{*}{\textit{Sherlock}} & WCP & 79.22 & 83.44 & 0.78 \\
 & NMI & 48.90 & 53.37 & 0.78 \\
\bottomrule
\end{tabular}}
\end{table}

\begin{figure*}[!t]
\centering
\includegraphics[width=0.95\textwidth]{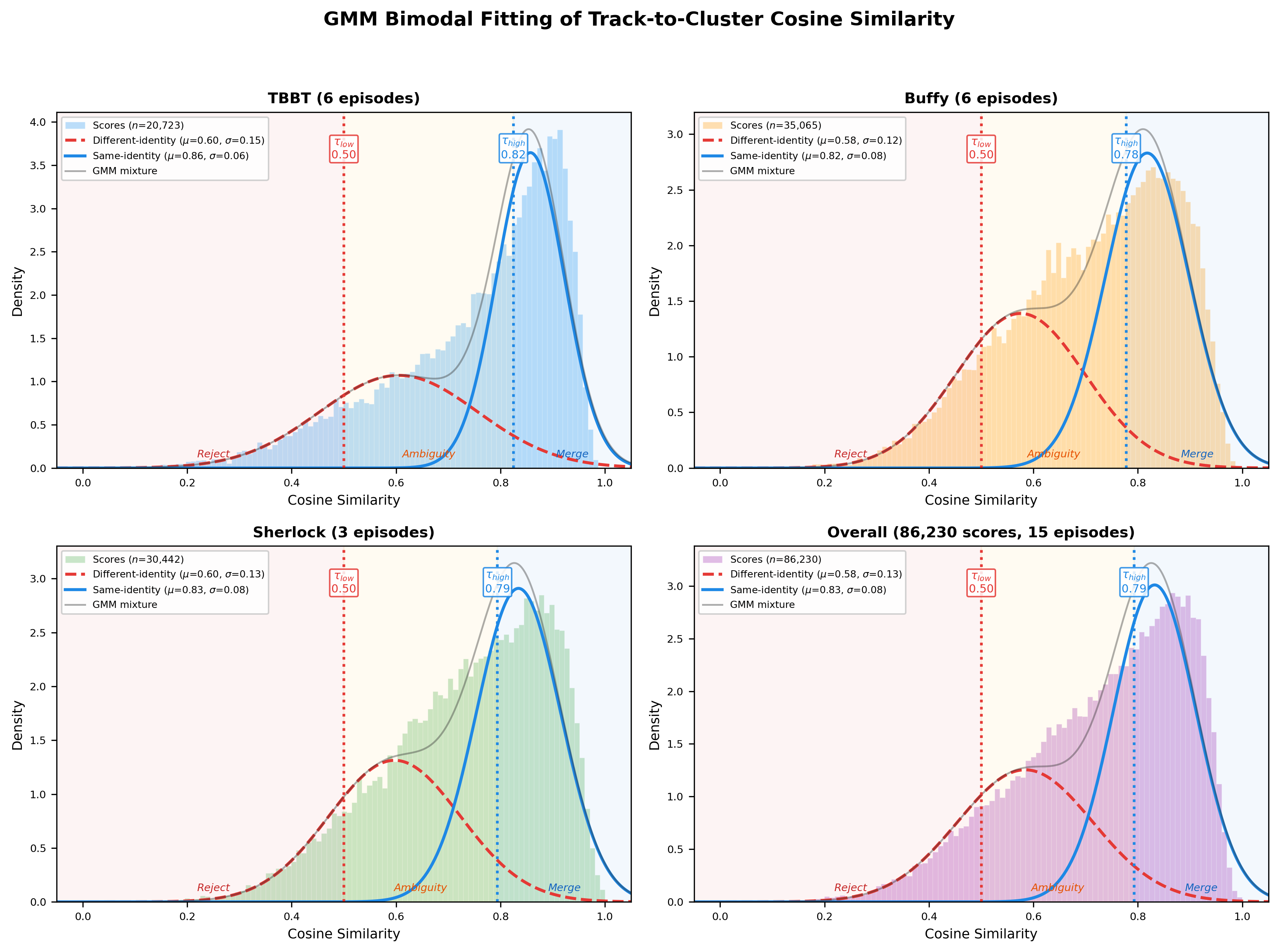}
\caption{GMM bimodal fitting of track to cluster cosine similarity distributions across all 15 episodes (86,230 total scores). Each panel shows the empirical histogram overlaid with the fitted two component GMM. The dashed red and solid blue curves represent the different identity and same identity components, respectively. Vertical dotted lines indicate the derived $\tau_{\text{low}}$ and $\tau_{\text{high}}$ thresholds after clipping.}
\label{fig:gmm_fitting}
\end{figure*}

\paragraph{Marginal effects.}
Table~\ref{tab:lambda_marginal} isolates the marginal effect of each specific semantic trait. We observe that persistent traits (gender and age) heavily dictate performance. Specifically, the gender mismatch penalty $\lambda_{\text{gm}}$ and gender+age reward $\lambda_{\text{ga}}$ exhibit the largest influence on the Balanced Average. In contrast, transient cues like clothing reward $\lambda_{\text{cloth}}$ and hair conflict $\lambda_{\text{hc}}$ are comparatively insensitive, validating VISTA's decision to decouple transient and persistent traits.

\begin{table}[ht]
\centering
\caption{Cross dataset marginal effects of arbitration weights. Each row averages over 81 configurations sharing that specific weight value. Bal.\ Avg.\ = (WCP + NMI + ID\_Acc) / 3.}
\label{tab:lambda_marginal}
{
\begin{tabular}{llccc}
\toprule
\multirow{2}{*}{Weight} & \multirow{2}{*}{Value} & \multicolumn{3}{c}{Bal.\ Avg.\ (\%)} \\
\cmidrule(lr){3-5}
 & & \textit{TBBT} & \textit{Buffy} & \textit{Sherlock} \\
\midrule
\multirow{3}{*}{$\lambda_{\text{ga}}$}
 & 0.04 & 93.88 & \textbf{83.53} & 65.22 \\
 & 0.08 & 93.81 & 82.42 & 65.70 \\
 & 0.12 & \textbf{94.01} & 82.40 & \textbf{66.29} \\
\midrule
\multirow{3}{*}{$\lambda_{\text{cloth}}$}
 & 0.02 & \textbf{93.93} & \textbf{82.82} & 65.63 \\
 & 0.05 & 93.88 & 82.81 & 65.76 \\
 & 0.08 & 93.89 & 82.72 & \textbf{65.82} \\
\midrule
\multirow{3}{*}{$\lambda_{\text{gm}}$}
 & 0.08 & \textbf{93.95} & 82.74 & 65.64 \\
 & 0.12 & 93.93 & 82.77 & 65.63 \\
 & 0.16 & 93.82 & \textbf{82.84} & \textbf{65.95} \\
\midrule
\multirow{3}{*}{$\lambda_{\text{am}}$}
 & 0.01 & \textbf{93.94} & 82.68 & \textbf{65.77} \\
 & 0.03 & 93.88 & 82.76 & 65.72 \\
 & 0.05 & 93.88 & \textbf{82.90} & 65.73 \\
\midrule
\multirow{3}{*}{$\lambda_{\text{hc}}$}
 & 0.02 & \textbf{93.93} & \textbf{82.90} & 65.73 \\
 & 0.04 & 93.88 & 82.73 & 65.72 \\
 & 0.06 & 93.89 & 82.72 & \textbf{65.77} \\
\bottomrule
\end{tabular}}
\end{table}

\paragraph{Robustness over absolute peaks.}
It is worth noting that the default weights used in the main paper's unified evaluation (e.g., $\lambda_{\text{gm}}=0.12, \lambda_{\text{cloth}}=0.04$) were deliberately selected as a robust global compromise rather than absolute dataset specific peaks. While Table~\ref{tab:lambda_marginal} shows that tuning these weights per dataset could yield marginally higher scores (e.g., \textit{Buffy} peaks at $\lambda_{\text{ga}}=0.04$ while \textit{Sherlock} peaks at $0.12$), the unified setting ensures strong out of the box generalization. The minimal performance variance across these settings ultimately proves that VISTA's effectiveness stems from its core architectural design symbolic denoising and spatio temporal triage rather than fragile hyperparameter overfitting.

\section{GMM Bimodal Fitting Visualization}
To empirically validate the bimodal assumption underlying VISTA's dynamic thresholding (Section 3.3 of the main paper), we visualize the GMM fit across all 15 episodes in Figure~\ref{fig:gmm_fitting}. The cosine similarity scores between track cluster pairs are aggregated per dataset and fitted with a two component GMM.

All three datasets exhibit clear bimodal structure, with the same identity component concentrated at $\mu \approx 0.82$-$0.86$ and the different identity component at $\mu \approx 0.58$-$0.60$. \textit{TBBT} shows the tightest same identity distribution ($\sigma{=}0.06$), consistent with its controlled studio lighting. \textit{Buffy} and \textit{Sherlock} exhibit wider spread ($\sigma{=}0.08$), reflecting greater illumination and appearance variation. The estimated $\tau_{\text{low}}$ and $\tau_{\text{high}}$ thresholds consistently partition the score space into well separated reject, ambiguity, and merge zones. The raw GMM derived $\tau_{\text{low}}$ values ($\mu_1 + \sigma_1 \approx 0.70$-$0.75$) are bounded via the online proposal rule $\tau_{\text{low}}^\star=\mathrm{clip}(\mu_1+\sigma_1,\,0.15,\,0.50)$, followed by EMA smoothing with $\eta_{\text{gmm}}=0.3$ (Section 3.3 of the main paper). The $0.50$ cap prevents the lower boundary from drifting too high in visually compressed windows, preserving an ambiguity zone in which semantic arbitration can operate; it is an intentional design choice rather than a fitting artifact. In an isolated online cap sweep over all 15 episodes, relaxing the cap to $0.70$ or $1.00$ yields identical WCP/NMI/ID\_Acc after rounding to two decimals: once the ambiguity zone is preserved, downstream best assignment selection and the arbitration gate $\tau_{\text{merge}}$ dominate the final accept/reject decisions. The cap therefore acts as a declared stability safeguard rather than a tuned performance knob.

\section{Intra Shot Tracklet Length Ablation}

Table~\ref{tab:supp_length_ablation} expands upon the tracklet length ablation discussed in Section 4.4 of the main paper. As shown, increasing the maximum intra shot tracklet length from 12 to 48 frames improves Identity Accuracy (ID\_Acc) and NMI by accumulating richer temporal context and producing more stable running average embeddings.

However, over extending the window to 96 frames begins to degrade performance due to intra track appearance noise (e.g., characters turning completely away from the camera). Note that Precision, Recall, and F1 remain perfectly constant across these variations. This is expected, as the upstream face detection bounding boxes and initial track candidate generation are fixed; the tracklet length parameter only dictates downstream cluster aggregation and inter shot tracking logic. Based on this trade off, we adopted $L_{\text{max}}=24$ as our unified default to maintain high Weighted Cluster Purity (WCP) without sacrificing latency.

\begin{table}[ht]
\centering
\caption{Intra shot tracklet length ablation, averaged over all 15 episodes (WCP / NMI / ID\_Acc / Precision / Recall / F1 \%). Upstream detection metrics (Precision, Recall, F1) are naturally unaffected by downstream aggregation logic.}
\label{tab:supp_length_ablation}
\resizebox{\columnwidth}{!}{
\begin{tabular}{lcccccc}
\toprule
Length & WCP & NMI & ID\_Acc & Precision & Recall & F1 \\
\midrule
12 & 92.27 & 77.39 & 77.88 & 81.16 & 96.19 & 87.84 \\
24 & \textbf{92.83} & 77.92 & 79.14 & 81.16 & 96.19 & 87.84 \\
48 & 91.89 & \textbf{78.72} & \textbf{81.05} & 81.16 & 96.19 & 87.84 \\
96 & 91.23 & 78.50 & 80.79 & 81.16 & 96.19 & 87.84 \\
\bottomrule
\end{tabular}}
\end{table}

\section{Additional Qualitative Results}
\label{sec:supp_qual}
\begin{figure*}[!t]
    \centering
    \includegraphics[width=0.95\textwidth]{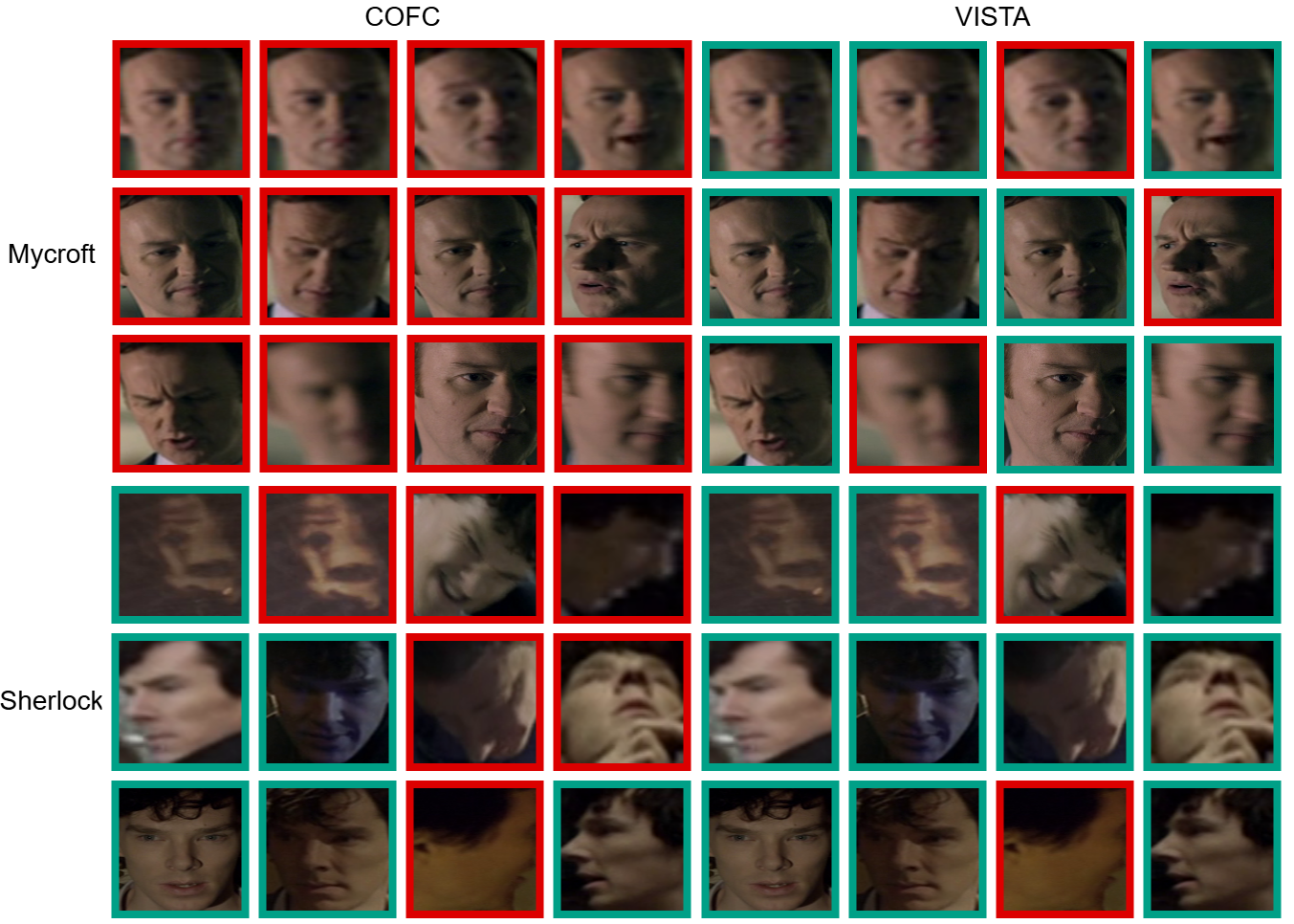}
    \caption{Local clustering purity for Mycroft and Sherlock (Green: true positives, Red: false positives). VISTA maintains higher purity than COFC across challenging pose, illumination, and motion blur variations, minimizing identity confusion.}
    \label{fig:supp_qual_grid}
\end{figure*}

Figure~\ref{fig:supp_qual_grid} provides the crop level qualitative comparison on \textit{Sherlock} referenced in Section 4.5 of the main paper (moved here for space). Under challenging pose, illumination, and motion blur variations, VISTA maintains higher local purity than COFC and minimizes identity confusion between visually similar characters.

\section{Additional Online Baselines: Full Results}
\label{sec:supp_baselines}
This section provides the full per dataset results for the two additional online baselines summarised in Section 4.3 of the main paper. Both baselines use the same detector (RetinaFace), embeddings (FaceNet), tracks, and evaluation protocol as VISTA and COFC.

\paragraph{Naive VLM hard constraint baseline.}
Raw VLM attributes (no SSR, no triage) are applied as must link/cannot link rules. Table~\ref{tab:supp_hardvlm} shows severe over fragmentation: the number of predicted clusters exceeds ground truth by $4.0$--$11.3\times$. This inflates WCP while ID\_Acc collapses far below VISTA Full ($93.94$/$82.82$/$68.10$ on \textit{TBBT}/\textit{Buffy}/\textit{Sherlock}) and below the weakest geometry driven ablation in the main paper, confirming that VISTA's gains stem from selective semantic triage rather than from VLM attributes alone.

\begin{table}[ht]
\centering

\caption{Naive VLM hard constraint baseline (\%). \#Pred/ep and \#GT/ep are per episode averages by series.}
\label{tab:supp_hardvlm}
\resizebox{\columnwidth}{!}{
\begin{tabular}{lccccc}
\toprule
Hard-VLM & WCP & NMI & ID\_Acc & \#Pred/ep & \#GT/ep \\
\midrule
\textit{TBBT}     & 97.98 & 76.57 & 63.95 & 83.8  & 20.5 \\
\textit{Buffy}    & 91.01 & 72.99 & 60.64 & 108.5 & 27.0 \\
\textit{Sherlock} & 84.67 & 44.98 & 40.65 & 136.0 & 12.0 \\
\bottomrule
\end{tabular}}
\end{table}

\paragraph{Sequential online HAC baseline.}
A sequential online HAC variant on the same tracks reaches very high WCP, as expected under fragmentation, but substantially lower NMI and ID\_Acc, with severe over fragmentation (Table~\ref{tab:supp_hac}).

\begin{table}[ht]
\centering

\caption{Sequential online HAC baseline (\%). The last column is the ratio of predicted clusters to ground truth clusters per episode.}
\label{tab:supp_hac}
\resizebox{\columnwidth}{!}{
\begin{tabular}{lcccc}
\toprule
Online HAC & WCP & NMI & ID\_Acc & \#Pred/\#GT \\
\midrule
\textit{TBBT}     & 99.73 & 84.93 & 81.29 & $4.5\times$ \\
\textit{Buffy}    & 97.62 & 76.78 & 63.49 & $9.5\times$ \\
\textit{Sherlock} & 91.19 & 49.28 & 40.56 & $39.8\times$ \\
\bottomrule
\end{tabular}}
\end{table}

\section{Additional Threshold Sensitivity Analyses}
\label{sec:threshold_grid}

\paragraph{Warm up threshold grid.}
Complementing the marginal tables in Section~S5, we run a dedicated $3\times3$ grid bracketing the warm up initialisation over $\tau_{\text{high}}^{(0)}\in\{0.75, 0.825, 0.90\}$ and $\tau_{\text{low}}^{(0)}\in\{0.20, 0.275, 0.35\}$ with all other settings fixed (135 episode level evaluations). Table~\ref{tab:supp_grid} reports the per dataset max$-$min variation, which remains below $0.7$\,pp for all metrics. $\tau_{\text{high}}$ shows limited sensitivity in this grid precisely because GMM dynamic thresholding overrides the static initialisation after warm up.

\begin{table}[ht]
\centering

\caption{Max$-$min variation (pp) across the $3\times3$ warm up threshold grid (all other settings fixed).}
\label{tab:supp_grid}
{
\begin{tabular}{lccc}
\toprule
Metric & \textit{TBBT} & \textit{Buffy} & \textit{Sherlock} \\
\midrule
WCP     & 0.18 & 0.15 & 0.67 \\
NMI     & 0.41 & 0.15 & 0.30 \\
ID\_Acc & 0.67 & 0.37 & 0.69 \\
\bottomrule
\end{tabular}}
\end{table}

\paragraph{$\tau_{\text{merge}}$ sensitivity.}
We sweep the arbitration acceptance gate over $\tau_{\text{merge}}\in\{0.280, 0.325, 0.375, 0.425, 0.475\}$. The sweep confirms the expected purity--continuity trade off: %
at the most conservative setting ($0.475$), \textit{Sherlock} ID\_Acc drops by $8.37$\,pp relative to the $0.325$ default, reflecting conservative fragmentation rather than a standalone visual threshold failure.

\paragraph{$\tau_{\text{low}}$ cap sweep.}
As reported in Section~S8, an isolated online cap sweep over all 15 episodes shows that relaxing the $0.50$ cap on $\tau_{\text{low}}^\star$ to $0.70$ or $1.00$ yields identical WCP/NMI/ID\_Acc after rounding to two decimals, confirming that the cap is a declared stability safeguard rather than a tuned performance knob.

\section{Subgroup Clustering Quality}
\label{sec:supp_fairness}
Complementing the triage rate analysis in the main paper (Section 4.7), Table~\ref{tab:supp_fairness} reports subgroup clustering quality gaps on the IoU matched face subset, using the same VLM predicted gender/age groups. VPCD provides no ground truth demographic annotations, so these figures should be read as a diagnostic rather than a full fairness audit. The signs of the gaps vary across datasets, indicating character and dataset dependent effects rather than a one directional degradation. As an ethical safeguard, our implementation provides an option to disable the gender/age attributes entirely, in which case arbitration falls back to transient cues and geometry only.

\clearpage
\begin{table}[htbp]
\centering

\caption{Subgroup clustering quality gaps (pp) on the IoU matched subset using VLM predicted demographic groups. Positive values indicate the first group scores higher.}
\label{tab:supp_fairness}
{
\footnotesize
\setlength{\tabcolsep}{3pt}
\begin{tabular}{@{}llccc@{}}
\toprule
Gap & Dataset & WCP & NMI & ID\_Acc \\
\midrule
\multirow{3}{*}{\shortstack[l]{Gender\\(M$-$F)}}
 & \textit{TBBT}     & $-0.5$ & $+13.1$ & $+1.7$ \\
 & \textit{Buffy}    & $+2.1$ & $+14.7$ & $+5.0$ \\
 & \textit{Sherlock} & $-6.6$ & $-19.8$ & $-1.4$ \\
\midrule
\multirow{3}{*}{\shortstack[l]{Age\\(adult$-$senior)}}
 & \textit{TBBT}     & $+2.0$ & $+12.0$ & $+13.4$ \\
 & \textit{Buffy}    & $-2.3$ & $+4.7$  & $-0.6$ \\
 & \textit{Sherlock} & $-6.6$ & $+13.4$ & $+1.1$ \\
\bottomrule
\end{tabular}}
\end{table}

\bibliographystyle{ACM-Reference-Format}
\bibliography{sample-base}